\documentclass[10pt,twocolumn,letterpaper]{article}

\usepackage{iccv}
\usepackage{times}
\usepackage{epsfig}
\usepackage{graphicx}
\usepackage{subcaption}     % subcaption
\usepackage{amsmath}
\usepackage{amssymb}
\usepackage{algorithm}
\usepackage{algpseudocode}
\usepackage{enumitem}
\usepackage[pagebackref=true,breaklinks=true,letterpaper=true,colorlinks,bookmarks=false]{hyperref}

\DeclareMathOperator*{\argmin}{arg\,min}

\newcommand{\method}{TACO}
\renewcommand{\paragraph}[1]{\noindent\vspace{-0.0em}\textbf{{#1}}}
\iccvfinalcopy % *** Uncomment this line for the final submission

\def\iccvPaperID{10610} % *** Enter the ICCV Paper ID here

\ificcvfinal\fi

\begin{document}

%%%%%%%%% TITLE
\title{\method{}:  Vision Models Can Be Efficiently Specialized  \\ via Few-Shot Task-Aware Compression}
{\small
\author{
Denis Kuznedelev $^\dag$ \\
{\small Skoltech \& Yandex}\\
{\tt\small Denis.Kuznedelev@skoltech.ru}
\and
Soroush Tabesh $^\dag$ \\
{\small IST Austria \& Sharif UT}\\
{\tt\small soroush.tabesh@ist.ac.at}
\and
Kimia Noorbakhsh $^\dag$ \\
{\small IST Austria \& Sharif UT}\\
{\tt\small kimia.noorbakhsh@ist.ac.at}
\and
Elias Frantar $^\dag$ \\
{\small IST Austria}\\
{\tt\small elias.frantar@ist.ac.at}
\and
Sara Beery \\
{\small Massachusetts Institute of Technology}\\
{\tt\small beery@mit.edu}
\and
Eldar Kurtic \\
{\small IST Austria}\\
{\tt\small eldar.kurtic@ist.ac.at}
\and
Dan Alistarh \\
{\small IST Austria \& Neural Magic}\\
{\tt\small dan.alistarh@ist.ac.at}
}
}

\maketitle
\def\thefootnote{$^\dag$}\footnotetext{Equal contribution. Contact: {\tt\footnotesize Denis.Kuznedelev@skoltech.ru}, {\tt\footnotesize dan.alistarh@ist.ac.at}.}\def\thefootnote{\arabic{footnote}}
\ificcvfinal\thispagestyle{empty}\fi

%%%%%%%%% ABSTRACT
\begin{abstract}
   Recent vision architectures and self-supervised training methods enable vision models that are extremely accurate and general, but come with massive parameter and computational costs. In practical settings, such as camera traps, users have limited resources, and may fine-tune a pretrained model on (often limited) data from a small set of specific categories of interest. These users may wish to make use of modern, highly-accurate models, but are often computationally constrained. To address this, we ask: \textbf{can we quickly compress large generalist models into accurate and efficient specialists?} 
For this, we propose a simple and versatile technique called \emph{Few-Shot Task-Aware Compression (TACO)}. 
Given a large vision model that is pretrained to be accurate on a \emph{broad task}, such as classification over ImageNet-22K, TACO produces a smaller model that is accurate on \emph{specialized tasks}, such as classification across vehicle types or animal species. 
Crucially, TACO works in \emph{few-shot fashion}, i.e. only a few task-specific samples are used, and the procedure has low computational overheads. 
We validate TACO on highly-accurate ResNet, ViT/DeiT, and ConvNeXt models, originally trained on ImageNet, LAION, or iNaturalist, which we specialize and compress to a diverse set of ``downstream'' subtasks. 
TACO can reduce the number of non-zero parameters in existing models by up to 20x relative to the original models, leading to inference speedups of up to 3$\times$,  while remaining accuracy-competitive with the uncompressed models on the specialized tasks. 
\end{abstract}

\vspace{-1em}
\section{Introduction}

\begin{figure}[!t]
    \begin{subfigure}{\linewidth}
        \centering
        \includegraphics[width=0.95\linewidth]{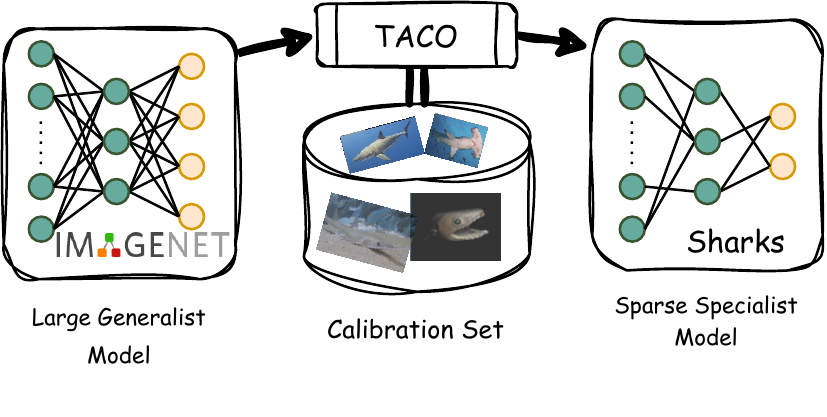}
    \end{subfigure}
    \vspace{-2em}
    \caption{
        Few-Shot Task-Aware COmpression (TACO) is done using a calibration set, which is simply a few-shot subset of the desired task. It consists of a single-step compression procedure, followed by a few-shot finetuning step. 
    }

    \label{fig:method_diagram}
\end{figure}

The recent introduction of new 
training techniques that can leverage massive amounts of unlabelled data, such as self-supervised (contrastive) learning~\cite{chen2020simple, radford2021learning}, complemented by new architectures, such as Vision Transformers (ViTs)~\cite{ViT, DeiT} and next-generation convolutional networks (e.g., ConvNeXt)~\cite{liu2022convnet}, has led to unprecedented accuracy on standard benchmarks, such as the ImageNet image classification task, even in zero- or few-shot settings. 

Yet, this impressive increase in accuracy comes at the price of large parameter counts and computational costs. For example, the highly accurate ``large'' variants of the Vision Transformer (ViT/DeiT) model family, commonly used for pretraining, can easily count more than 200 Million parameters, requiring more than 50 GFLOPs for inference at standard input sizes. Thus, although these models reach extremely high accuracies for image classification---e.g., close to 90\% Top-1 accuracy on ImageNet-1K---they are impractical to deploy, especially in resource-constrained settings, such as edge inference.  

One popular way of bridging this computational divide is \emph{post-training compression}, e.g.~\cite{nagel2020up, hubara2021accelerated, frantar-obc}:
given a small but representative calibration dataset, the model is compressed to a given sparsity or low-precision requirement \emph{in one compression step}. 
However, highly-accurate models, such as the ones from the Vision Transformer family, are notoriously hard to compress~\cite{SVIT}: intuitively, these models may be inherently parameter-heavy, as they are \emph{generalists}, trained across massive and often noisy datasets, and classifying images across thousands of output classes.  

In this paper, we seek to extend the reach of post-training compression to consider the requirement of \emph{model specialization}. 
We start from the idea that, in practical deployment scenarios, such as an ecologist seeking to identify animal species in camera trap images~\cite{beery2018recognition} or a city planner seeking to identify vehicles using a traffic camera~\cite{ozkurt2009automatic}, users only need their models to categorize a smaller set of ``classes of interest.'' Thus, we shift the objective to producing  \emph{efficient specialist models}, which maintain high accuracy on a narrower subset of the data distribution. 
Specialization should allow for higher compression, as only a fraction of the features should be required over narrower tasks. 

We propose a simple and versatile approach which we call \textbf{T}ask-\textbf{A}ware \textbf{CO}mpression (\method{}). 
(Please see Figure~\ref{fig:method_diagram} for an illustration.) 
Given a \emph{pre-trained generalist model} and a ``calibration set'' containing \emph{task-specific data},  \method{} can produce an efficient specialist model in two steps. 
The first is a \emph{single task-aware compression step}, which reduces the model size layer-wise, preserving the model's features on the task-specific calibration data, but removing unnecessary parameters. 
Here, our key observation is that some of the existing highly-accurate solvers for sparsity can also be leveraged for \emph{model specialization} over a task-specific calibration set. 
The second step is a \emph{few-shot tuning} procedure, which allows the model to recover accuracy by optimizing it globally. 
Here, our contribution is a lightweight tuning technique which does not overfit over small calibration sets.

We validate TACO experimentally on a wide range of pretraining settings, model architectures, and target tasks. 
Specifically, we start from  models pre-trained via self-supervised learning (using CLIP~\cite{radford2021learning} or DINO~\cite{DINO} on e.g. the LAION dataset~\cite{schuhmann2021laion}) or via standard supervised learning on the ImageNet ILSVRC12~\cite{ILSVRC15} and iNaturalist 2021~\cite{van2018inaturalist} datasets. 
We consider highly-accurate base models on these datasets from the ResNet~\cite{he2016deep}, Vision Transformer (ViT)~\cite{ViT, DeiT, DeiT3} and ConvNeXt~\cite{liu2022convnet} model families, which we specialize across tasks obtained by restricting the ImageNet and iNaturalist taxonomies to natural subsets, such as animal species or vehicle classes. 
In addition, we extend TACO to a standard transfer learning setting, showing that it is superior to alternatives such as Lottery-Ticket-based pruning~\cite{chen2021lottery}. 
We compress models with both unstructured and structured pruning, and examine model specialization, as well as computational gains vs. accuracy loss relative to the uncompressed models. 

%%%%%%%%% BODY TEXT
% \section{Introduction}

Our results show that, for unstructured pruning, TACO can reduce the number of model parameters by up to 10x without accuracy loss, and 20x with moderate accuracy loss, in the order of a few percentage points on the target task, relative to the original model. Our analysis shows that this occurs via \emph{task specialization}: TACO models have significantly better accuracy at the same compression level than models compressed via standard post-training compression, doubling compression on average. This behavior is preserved across model classes, tasks,  and sparsity levels. However, our analysis shows that residual networks (ResNet and ConvNeXt) are more compressible, as well as the fact that ``simpler'' specialized tasks, with fewer target classes, allow for higher compression due to specialization. 

The TACO procedure is both general and computationally efficient: for instance, we can specialize an 86M parameter ViT-Base model in 20 minutes on a single GPU. This allows us to perform a full ablation, specializing over a large fraction of the WordNet taxonomy (166 tasks) in reasonable time.
Moreover, TACO supports arbitrary solvers for imposing compression: we instantiate it to structured and unstructured pruning, and propose a ``hybrid'' solver, which is both accurate and highly-efficient.  
In terms of practical gains, we show that TACO models provide end-to-end computational speedups of up to 2x relative to the uncompressed models, at similar accuracy levels on both CPUs and GPUs, and 3x at moderate accuracy loss.  

% \section{Background and Related Work}

\section{Few-Shot Task-Aware Compression (TACO)}

\subsection{Task-Aware Post-Training Compression} 
%This is the ONE-SHOT section

The first step of the TACO procedure performs one-shot compression on the given model and task, reducing the model size in a layer-wise fashion by leveraging a set of task-specific calibration data $\mathcal{D}$. 
At a high level, given an uncompressed model $f$, a sparsity solver $S$, task-specific calibration data $\mathcal{D}$, and compression constraint $\mathcal{C}$, the TACO procedure
produces a compressed version $$f_\mathcal{C}^\mathcal{D} = \mathrm{TACO}(f, S, \mathcal{C}, \mathcal{D}).$$

To implement this, we follow the post-training compression approach: we split the network layer-wise, and run task-specific calibration data $\mathcal{D}$ through the uncompressed model. 
For each network layer $\ell \in \{1, 2, \ldots, L\}$ with weights $\mathbf{W}_\ell$, we extract the input $\mathbf{X}_\ell^{\mathcal{D}}$ on the calibration data $\mathcal{D}$. 
Compression is then applied to each layer by solving a layer-wise optimization problem, i.e. identifying the layer weights $\widehat{\mathbf{W}}$ which minimize the difference between the dense and compressed layer output on the provided calibration data, measured in L2-norm. 
For example, for a linear layer, our goal is to determine a set of weights $\widehat{\mathbf{W}}_\ell^{\star}$ satisfying a  compression predicate $\mathcal{C}$ such that: 

\begin{equation}\label{eq:main}
\widehat{\mathbf{W}}_\ell^{\star} = \argmin_{\widehat{\mathbf{W}}_\ell \in \mathcal{C}} \Vert \widehat{\mathbf{W}}_\ell \mathbf{X}_\ell^{\mathcal{D}} - \mathbf{W}_\ell \mathbf{X}_\ell^{\mathcal{D}} \Vert_2^{2}.
\end{equation}

The key difference relative to prior work on post-training compression is \emph{specialization over task-aware calibration data}: 
while prior work uses \emph{generic} calibration data, our hypothesis is that, when applied over task-specific calibration data, a good solver for this constrained optimization problem can 
also \emph{isolate the features or parameters} that are specific to the specialized task. This should be done by considering the task-specific calibration data during the solving, as the weights alone are initially generalists.

% \paragraph{Example: ImageNet Sharks.} 
% As a simple illustration, we can compare the performance of 

% examine the impact of solving this compression problem over a general calibration set, relative to 

\vspace{-0.5em}
\subsubsection{Solving the Task-Aware Compression Problem}
\label{sec:solvers}

\paragraph{Sparsity Solvers.} 
To test the specialization hypothesis, we investigated the four existing solvers for unstructured sparsity in the post-training setting:

\begin{enumerate}[noitemsep]
    \item \emph{Magnitude Pruning}~\cite{han2015deep}, which sorts the weights by their absolute value, is \emph{data-agnostic}, so we use it as a reasonable baseline.  
    \item \emph{AdaPrune}~\cite{hubara2021accelerated} is a solver which first sets a mask via magnitude, and then optimizes the remaining weights via SGD \emph{over the calibration data}. This solver can be data-aware by optimizing over task-specific data.  
    \item The \emph{Optimal Brain Compression (OBC)} solver~\cite{frantar-obc} approximates an optimal solution for the problem in Equation~\ref{eq:main} by greedily pruning the weight with the lowest impact on the layer's L2 loss, and updating all remaining weights to best compensate for the incurred compression error. The solver's asymptotic complexity is $\mathcal{O}(d_{row} \cdot d_{col}^{3})$, where $d_{row}$ is the output dimension of the layer and $d_{col}$ 
    is either the number of input features (for a linear layer) or the number of channels multiplied by kernel size for a convolutional layer.
    This is the most accurate known solver for layer-wise sparsity but can take several hours to apply to a model with more than 100M parameters. 
    
    \item The \emph{FastOBC} (SparseGPT)~\cite{sparseGPT} solver approximates the OBC approach by pruning in a computationally-efficient block pattern. The solver is $> 100$x faster than OBC for the same layer size, at the cost of lower accuracy in the provided solution at a given sparsity.   
    
\end{enumerate}

\paragraph{A Hybrid Solver.} We also propose \emph{HybridOBC}, a solver which can be seen as a best-of-both-worlds composition of OBC and FastOBC. 
One can observe that most of the computational complexity of OBC $\mathcal{O}(d_{row} \cdot d_{col}^{3})$ comes from the layers with largest input dimensions.
Therefore, we propose to prune layers with smaller input dimensions with the more accurate OBC pruner, and the ones with large values of $d_{col}$
(for example, \texttt{fc2} layer in ViTs) with FastOBC. 
We will show that HybridOBC significantly reduces the runtime cost relative to OBC, with negligible loss of accuracy.

A full ablation with respect to these solvers (with and without TACO tuning) is presented in Section~\ref{sec:solvers}. 

% \subsubsection{Case Study: ImageNet Sharks}

% To illustrate the difference between different solvers, we consider the toy example of specializing and compressing a ResNet50 model pre-trained over ImageNet-1K over the subset of ImageNet shark classes. The experiments are performed in few-shot, across a sparsity range from 50\% to 95\%, applied uniformly per layer. The results are presented in Figure~\ref{fig:sharks_comparison}. 

% \dan{describe and analyze results of the experiment on Sharks}

% \begin{figure}[!h]
%     \begin{subfigure}{\linewidth}
%         \includegraphics[width=\linewidth]{figures/sharks/resnet50-sharks-methods.pdf}
%     \end{subfigure}

%     \caption{
%         ResNet50, Sharks, All sparsities, solver ablation \soroush{caption needed}
%         %\soroush{I feel the line plot doesn't show the success of the method:)) bar plot, maybe?} \sara{I also find it hard to compare the correct lines back and forth between the two. Maybe a grid of 4 small plots, one for each sparsity method? though that seems like wasted space maybe. Maybe a single plot with error difference for the two methods for each level of sparsity?}
%     }
%     \label{fig:sharks_comparison}
%     \vspace{-1em}
% \end{figure}

\vspace{-0.5em}
\subsubsection{Structured Pruning} 
\vspace{-0.5em}

\

So far, we have discussed compression via fine-grained weight pruning. 
However, the same approach works for imposing other types of compression: 
We also investigate structured pruning (i.e., removing rows/columns or filters from the weight matrices), for which we employ the state-of-the-art ZipLM~\cite{kurtic2023ziplm}  layer-wise constraint solver. 
ZipLM determines the least important input channels based on their saliency with respect to L2 loss on calibration data.

\subsection{Few-Shot Task-Aware Tuning} 
% \subsection{Efficient Performance Recovery}
At high compression rates ($>$ 10x), the above single-step compression approach can lead to significant performance degradation. 
A natural follow-up, for instance, in the few-shot setting~\cite{DINO}, is fine-tuning the model on a small amount of task-specialized data. However, doing this directly runs into a series of difficulties. 
First, we only have a limited amount of data (e.g., 5 samples per class), and we found that standard training can lead to ``overfitting'' over the calibration data. (That is, the calibration samples are ``memorized,'' but validation performance does not improve.) 
Second, we are finetuning \emph{sparse} models, which tend to be much harder to train~\cite{evci2019difficulty}. 
We address this by introducing a self-distillation-based approach~\cite{zhang2019your,zhang2021self} to recover the accuracy of the sparsified model relative to the dense one, which does not overfit when applied over a small calibration set. 

Specifically, our approach, called TACO tuning, minimizes the $L_2$ distance between the dense and sparse model logits, over the calibration set: 

\begin{equation}\label{eq:self_distillation_definition}
\begin{split}
\argmin_{\widehat{\mathbf{W}}_s} \Vert f_{\widehat{\mathbf{W}}_s}(\mathcal{D}) - f_{\mathbf{W}_d}(\mathcal{D}) \Vert_2^{2} 
\\ s.t. \quad (\mathbf{1}-\mathcal{M}_{{\mathbf{W}}_s}) \odot  \widehat{\mathbf{W}}_s = \mathbf{0}, 
\end{split}
\end{equation}
\\
where $f_\mathbf{W}$ denotes the model with parameters $\mathbf{W}$ and $\mathcal{M}_\mathbf{W}$ its corresponding sparsity mask. $\mathbf{W_s}$ and $\mathbf{W_d}$ represent the parameters of the sparse and the dense model respectively and $\mathcal{D}$ is the calibration set.

We have found this approach to have a number of advantages relative to standard supervised training or other distillation objectives: first, the method achieves good accuracy recovery across all tasks without the need for any hyper-parameter tuning; second, the use of L2 loss appears to alleviate the ``vanishing gradient'' problem which occurs when fine-tuning sparse models via e.g., cross-entropy loss; third, the use of self-distillation prevents overfitting. 

\section{Experimental Evaluation}

\subsection{Experimental Setup} 

We validate our approach across a diverse series of models and image classification tasks. 

\paragraph{Generalist Datasets.} Specifically, we consider two large image classification benchmarks: 
ImageNet ILSVRC12~\cite{ILSVRC15}, and iNaturalist 2021~\cite{van2018inaturalist}. 
For the former, we consider three categories of training setups for producing uncompressed models: 
1) unsupervised pretraining on a large corpus, e.g. LAION~\cite{schuhmann2021laion}, followed by ImageNet-1K finetuning; 
2) large-scale supervised pretraining on ImageNet-22K followed by  ImageNet-1K finetuning; 
3) standard ImageNet-1K pretraining. 
For iNaturalist, we consider a Masked AutoEncoder (MAE) \cite{he2022masked} pre-trained model and finetune it
following the same procedure and hyperparameters as the original paper. 

\paragraph{Generalist Models and Methods.} 
We consider three model families: residual CNNs (ResNets)~\cite{he2016deep}, 
data-efficient vision transformers (ViT/DeiT)~\cite{ViT,DeiT,DeiT3}, 
and modern transformer-inspired CNNs (ConvNeXt)~\cite{liu2022convnet}. 
We compress models across scales, from the relatively small
DeiT-S/16(224) (22M parameters) to 
relatively large vision models such as DeiT-H/14(224) (600M parameters). 

In experiments, we denote \emph{general post-training compression methods}, which are not task-aware, as \textbf{PTC}.   
These include post-training weight pruning, or structured pruning over a generic calibration set. 

\paragraph{Task Specialization.} 
For ImageNet experiments, we consider specialization across subtasks  of the WordNet hierarchy 
corresponding to ImageNet-1K. We generated those subtasks via the standard \texttt{robustness} library~\cite{robustness}. 
We selected subtasks that contain at least $5$ classes: this totals 166 possible subtasks on ImageNet. 
For iNaturalist, we specialize on various branches of the taxonomic tree corresponding to the following levels, in increasing order of specificity: kingdom, phylum, order, and genus. 

\paragraph{TACO Single-Step Pruning and Linear Probing.}
For single-step pruning, unless otherwise stated, we employ our HybridOBC solver, with a \emph{uniform} unstructured sparsity constraint across all linear and convolutional layers. (For ConvNeXt, we do not prune the depth-wise convolutions, as they have small computational and parameter impact.) 
Non-uniform sparsity distributions can lead to better accuracy, but may reduce end-to-end speedups--we leave an investigation of non-uniform sparsities for future work. 
We construct the subset-specialized classifier from the generalist classifier by 
keeping only the output channels relevant to the subset of interest in the output layer.
In order to recover model performance we apply linear probing on top of a sparse frozen 
backbone. Our linear probing setup is described in detail in the Appendix.
The TACO tuning procedure takes $\sim$ 10 minutes on a single NVIDIA A6000 GPU.

\subsection{Task-Agnostic vs. Task-Aware Compression}

Our first goal is to determine whether \emph{task-aware compression} 
can consistently lead to a better compression-accuracy trade-off on a given specialized task, relative to \emph{general} post-training compression (PTC). For this, a natural experiment is to apply TACO on a given sub-task (e.g., classifying vehicles on ImageNet), and compare the accuracy of TACO versus that of a general post-training compressor (using generic calibration data) both on the target task of interest (vehicles) but also on the generalist ImageNet validation set. 
We perform this experiment for various models, tasks, and compression objectives.

\begin{figure}[!h]
    \begin{subfigure}{\linewidth}
        \centering
        \includegraphics[width=\linewidth]{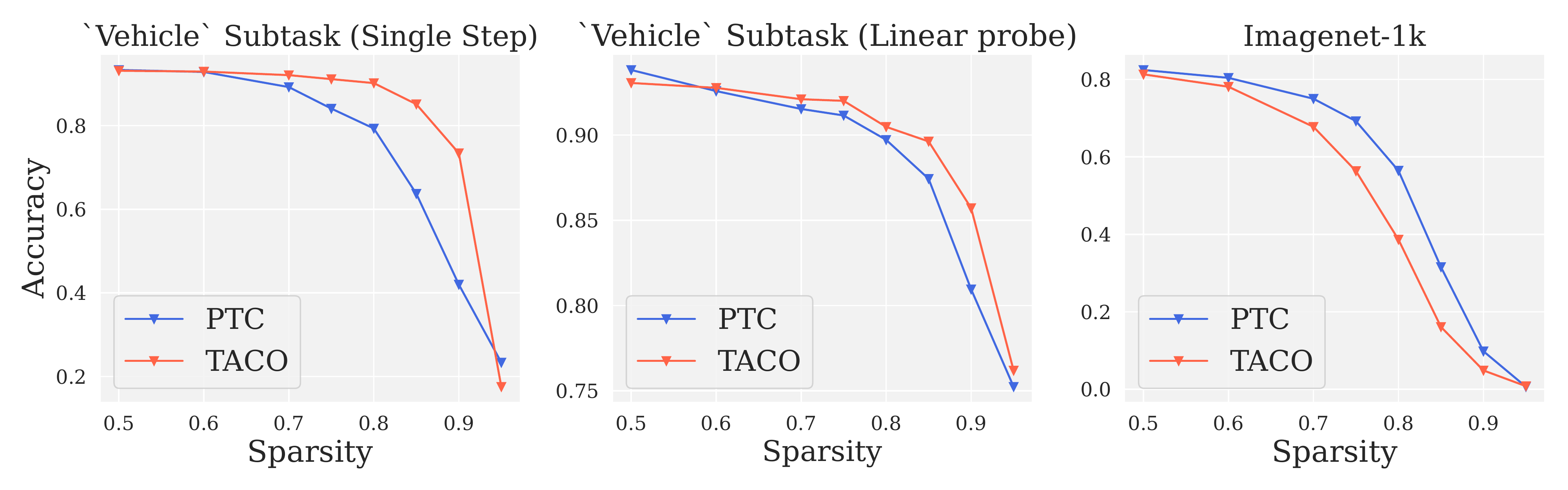}
    \end{subfigure}
    \begin{subfigure}{\linewidth}
        \centering
        \includegraphics[width=\linewidth]{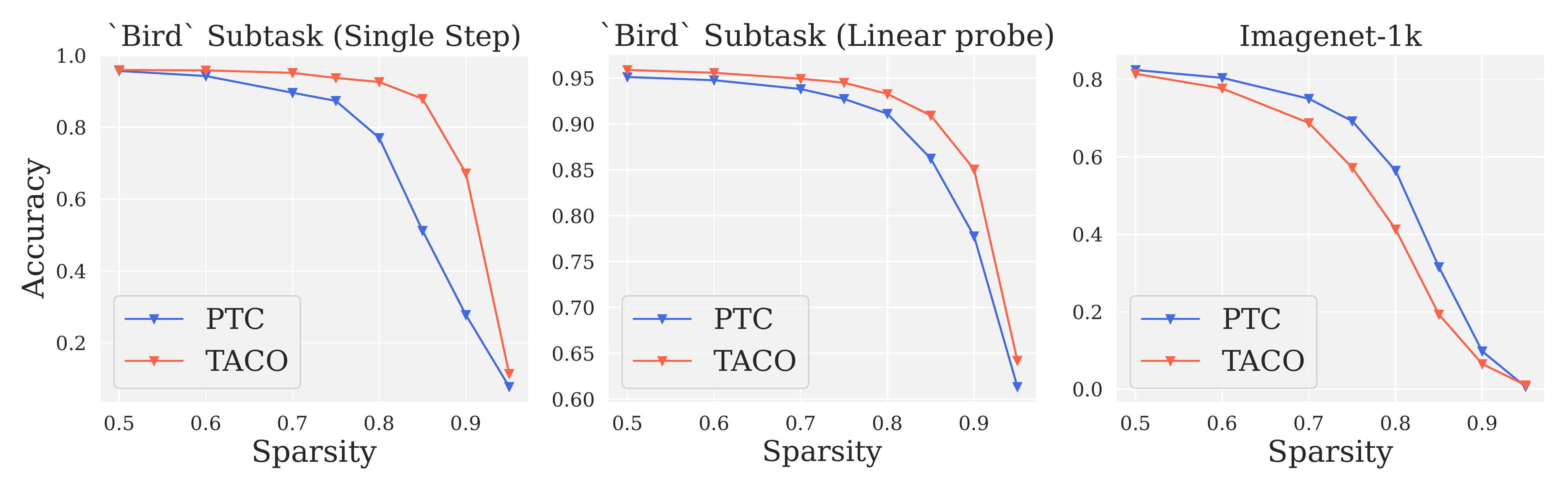}
    \end{subfigure}
    \caption{
        Unstructured pruning of the DeiT-III (B/16) model.
        Validation accuracy on a subtask (Vehicles, Birds) under various sparsities with
        (\textbf{Left}) single step compression and (\textbf{Middle}) linear probing. 
        (\textbf{Right}) Accuracy on the full ImageNet-1K validation set.
    }
    \label{fig:unstructured_pruning_ft}
    \vspace{-1em}
\end{figure}

\paragraph{Task-Aware Pruning of ImageNet Models.} 
First, we consider the specialization of models pre-trained on ImageNet-1k, for unstructured pruning,
 both via one-shot and TACO with finetuning. 
Figure \ref{fig:unstructured_pruning_ft} presents the accuracy results both on the target task (e.g., vehicles) 
and on the full Imagenet validation data.

% We take a model, prune it up to some specific sparsity ratio, and then train a linear classifier on top of it.
% At low sparsity levels, the accuracy is quite high even in a one-shot scenario. With the increase in sparsity,
% the model's performance starts to collapse, but even for high compression rates 
% (90\% - 95\%) one can recover quite a lot with the help of linear probing. 

We observe the following: 
First, \emph{task-aware} pruning performs better on the target task relative to  
 \emph{task-agnostic} pruning. Moreover, importantly, task-aware pruning can achieve significantly higher sparsity (e.g. $\geq 80\%$), at moderate accuracy loss, relative to general post-training compression, which sees notable drops in accuracy even at 60\% sparsity. Second, finetuning helps both approaches, 
 but preserves the advantage of the task-aware method, suggesting that feature specialization is occurring jointly with compression. 
 This observation is confirmed by the full ImageNet validation results, which show that the specialized model has lower performance on generic data relative to the generically-compressed model. 
In the Appendix, we present additional experiments across target tasks, showing that these findings hold very generally. 

% here is that the \emph{task-aware} pruner performs better on the given subset 
% of data compared to \emph{task-agnostic} pruner at the cost of lower performance on a more 
% general task (whole ImageNet). We evaluated the SparseGPT
% pruner on several subsets chosen at random. Results are presented on
% Figure \ref{fig:unstructured_pruning_ft}. 

\paragraph{Task-Aware Pruning of Self-Supervised Models.}  
Next, we consider models pre-trained with self-supervision (DINO) 
and image-language contrastive objectives (CLIP) that were not yet finetuned on the ImageNet dataset. 
We consider the ViT-B/32(224) and 
DINO-S/16(224) models. Since these models were not trained explicitly to predict 
ImageNet classes we train linear classifier on top of the vision model.
The linear probing setup and hyperparameters are the same as in the previous section. 
To evaluate performance, we prune model backbone and
perform linear probing on top of the frozen sparse feature extractor. 

The observation from the first 
experiment holds in this setting as well. 
Using calibration data from the target subset, one obtains a model that achieves higher accuracy 
on the expert task for the same compression level at the cost of higher drops on the general task. 
The behavior is similar for CLIP and DINO pre-trained models.

\begin{figure}[!h]
    \begin{subfigure}{\linewidth}
        \centering
        \includegraphics[width=\linewidth]{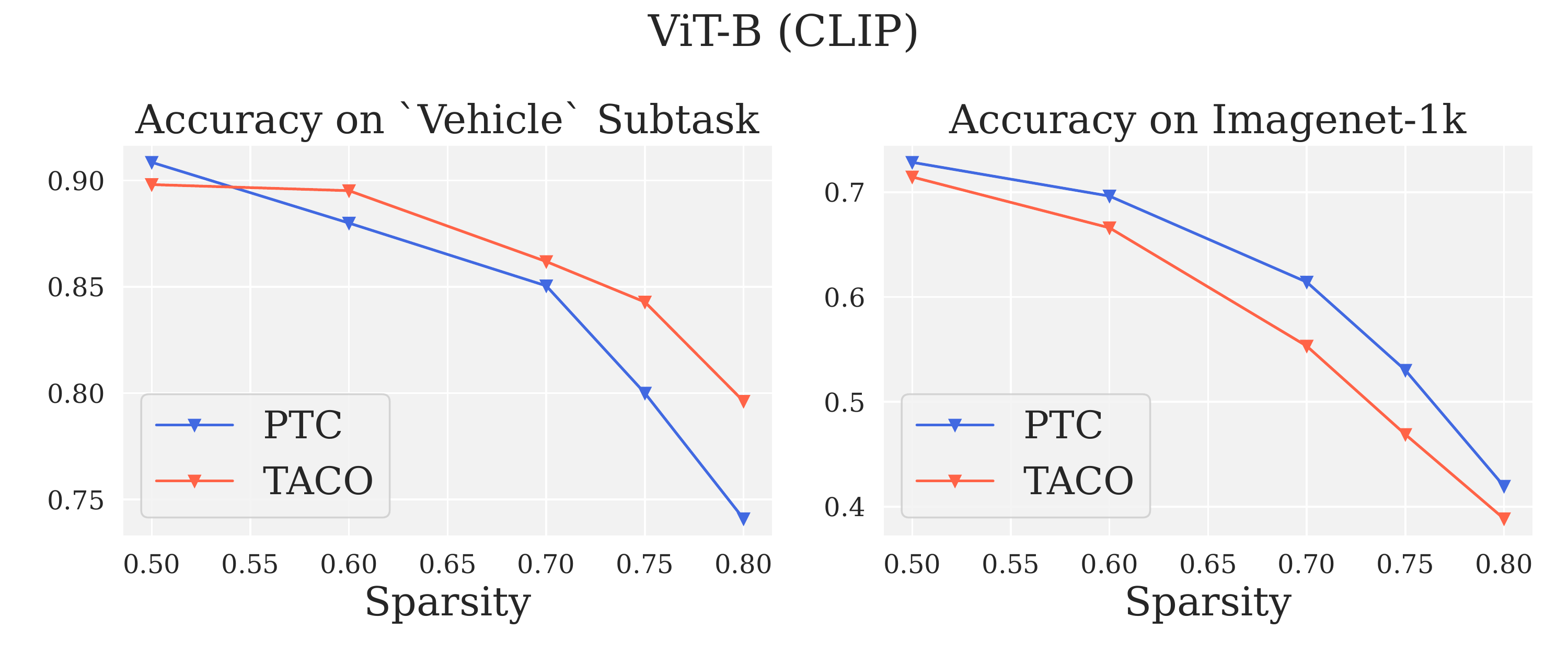}
    \end{subfigure}
    \begin{subfigure}{\linewidth}
        \centering
        \includegraphics[width=\linewidth]{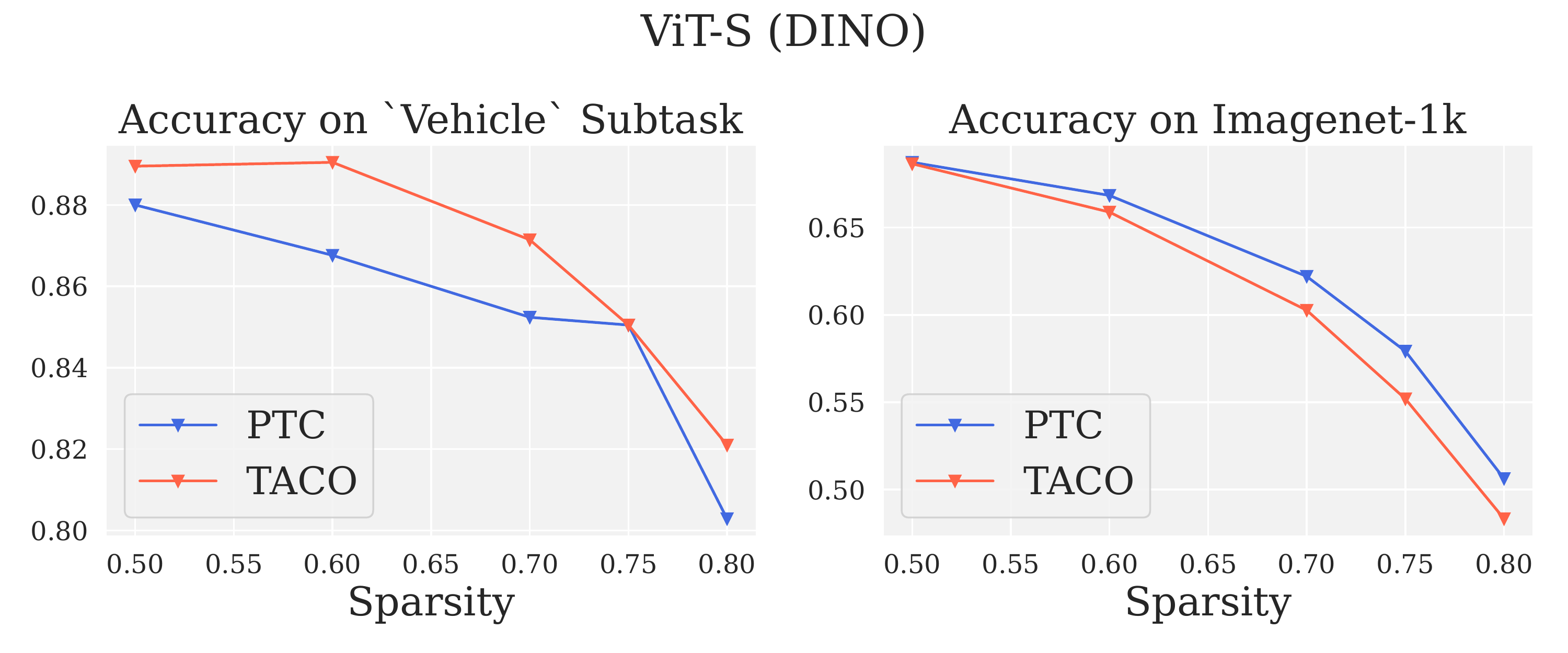}
    \end{subfigure}
    \caption{
        Unstructured pruning of self-supervised models, without ImageNet finetuning.
        (\textbf{Left}) Accuracy on the specialized target task (ImageNet subset). 
        (\textbf{Right}) Accuracy on the standard ImageNet-1K validation set.
    }
    \label{fig:unstructured_pruning_no_ft}
    \vspace{-1em}
\end{figure}

\newpage

\paragraph{CPU Inference Speedups.}
We next examine the practical potential of our method for CPU inference.
We use the DeepSparse runtime~\cite{deepsparse}, 
which provides inference acceleration of unstructured sparsity.
We consider the standard ResNet50 model, which has a relatively lower computational cost 
and is well-suited for edge inference. 
We prune the network via single-step TACO (with the HybridOBC solver) 
and then apply linear probing following the standard setup. 
Throughput measurements were conducted on an AMD EPYC 7452 Processor, using 4 cores.
 
\begin{figure}[!h]
    \centering
    \begin{subfigure}{\linewidth}
        \includegraphics[width=\linewidth]{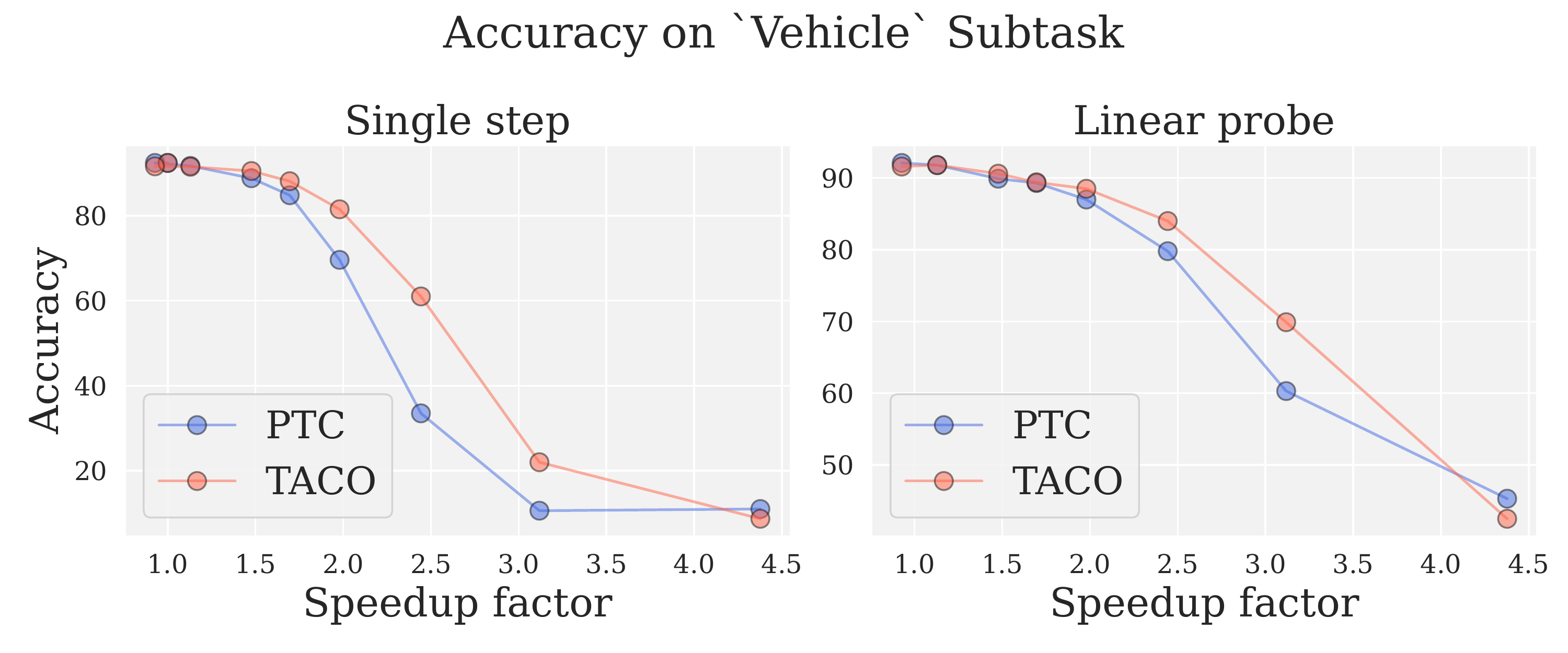}
    \end{subfigure}
    \caption{
        CPU speedup on ResNet50 relative to the dense model vs. accuracy for unstructured compression.
    }
    \label{fig:cpu_speedup}
    \vspace{-1em}
\end{figure}

\newpage

Figure~\ref{fig:cpu_speedup} presents the accuracy-vs-speedup trade-off for a ResNet50 model, 
specialized on the Vehicle subtask, for TACO with and without finetuning. 
One can achieve a speedup of up to 2.5x while still maintaining $\geq$ 80\% accuracy with TACO  
followed by tuning, and 2x using just TACO, without any finetuning.   

\subsection{The Impact of the Solver and Data}
\label{sec:exp_solvers}

\begin{figure}[!h]
    \centering 
    \begin{subfigure}{0.66\linewidth}
        \includegraphics[width=\linewidth]{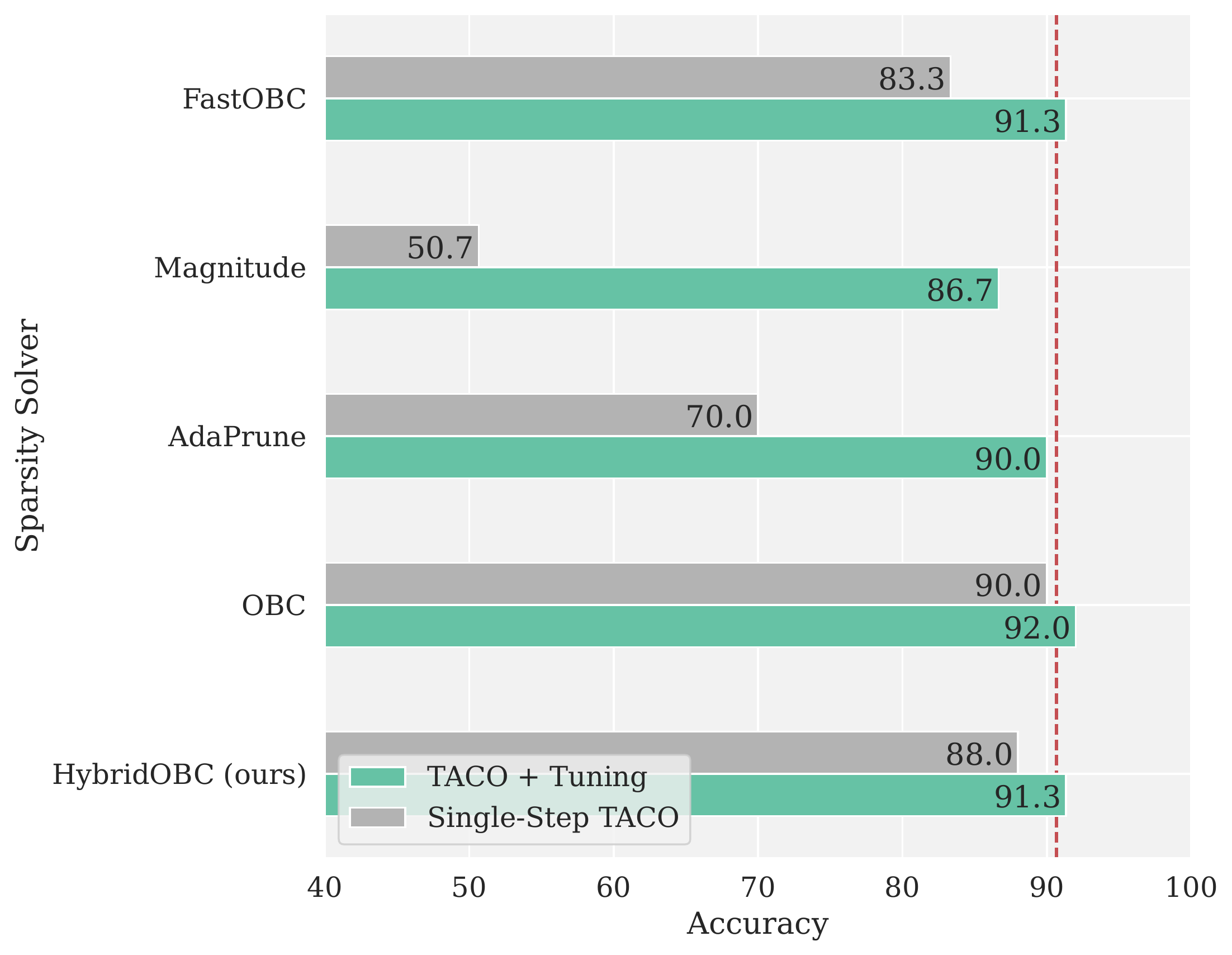}
    \end{subfigure}

    \caption{
        Comparison of solvers for task-aware compression of the ResNet50 model on the Sharks dataset, at 80\% sparsity, without (gray) and with finetuning (green). 
        The dense baseline is depicted as a dotted red line. 
    }
    \label{fig:sharks_solver_comparison}
    \vspace{-1em}
\end{figure}

\paragraph{Sparsity Solvers.} 
Next, we examine the impact of the choice of sparsity solver on specialization and accuracy. 
We consider unstructured pruning, and examine the solvers described in detail in Section~\ref{sec:solvers}: 
magnitude pruning, AdaPrune, OBC, FastOBC, and our HybridOBC proposal. 
Figure~\ref{fig:sharks_solver_comparison} compares the accuracy of these methods when specializing the ResNet50 model over the \texttt{sharks} target task, with and without finetuning. By contrast, PTC achieves significantly lower accuracy at the same speedup level. 

% Figure~\ref{fig:base_pruner_comparsion} shows a comparison of the accuracy of these methods when specializing the \texttt{deit3\_base\_patch16\_224} model over the Decapod target dataset, with and without finetuning. 

% \begin{figure}[!h]
%     \begin{subfigure}{\linewidth}
%         \centering
%         \includegraphics[width=\linewidth]{figures/pruner_comparison/deit3_base_patch16_224_decapod.pdf}
%     \end{subfigure}
%     \caption{
%          Comparison between sparsity solvers on DeiT-III (B/16) model.
%         (\textbf{Left}) Accuracy before finetuning. 
%         (\textbf{Right}) Accuracy after finetuning.
%         \denis{OBC is in progress}
%     }
%     \label{fig:base_pruner_comparsion}
%     \vspace{-1em}
% \end{figure}

We notice that OBC and HybridOBC are the most accurate solvers; 
they are followed by FastOBC, AdaPrune, and Magnitude Pruning. 
This order is intuitive: however, it is remarkable that OBC and HybridOBC provide almost the same accuracy. 
In terms of runtime, the compression step takes 4 minutes for FastOBC, 21 minutes for HybridOBC, and 90 minutes for OBC. Therefore, our HybridOBC proposal provides a good balance between accuracy and speed. 

\newpage

\paragraph{Filter Analysis.} 
Analyzing these results in more depth, we observe an interesting specialization: 
TACO almost completely removes weights corresponding to the Red color channel in the case of the `Sharks' task, which is natural since this color is less present in underwater images, and thus this channel is unlikely to provide useful information.  

\begin{figure}[!h]
    \centering 
    \begin{subfigure}{0.7\linewidth}
        \includegraphics[width=\linewidth]{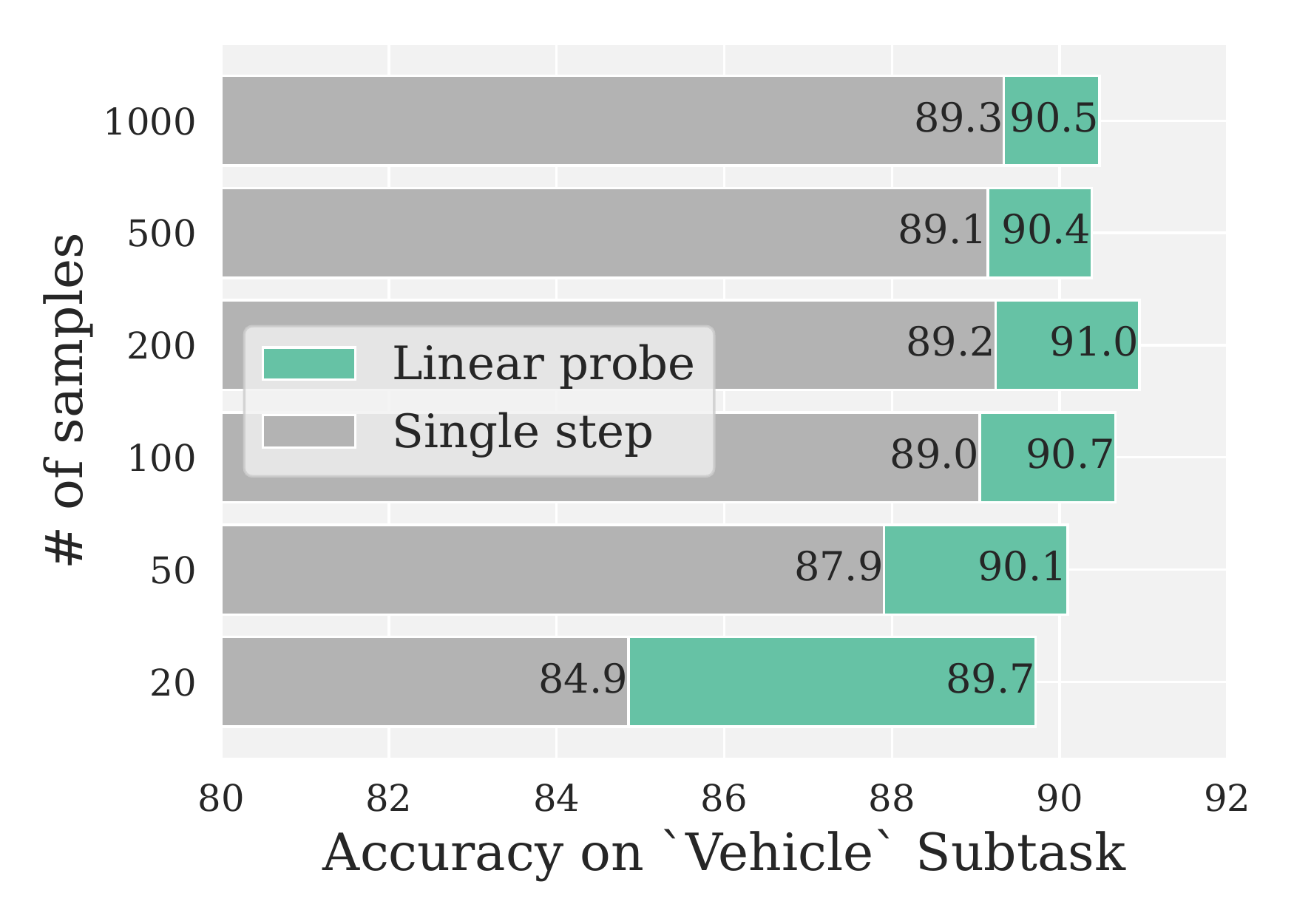}
    \end{subfigure}

    \caption{
        Accuracy versus calibration set size, at 80\% sparsity, without (gray) and with finetuning (green). 
    }
    \label{fig:calibration_comparison}
    \vspace{-1em}
\end{figure}

\paragraph{Calibration Set Size.}
In Figure~\ref{fig:calibration_comparison}, we examine the impact of the calibration set size on accuracy. 
We illustrate on the Vehicle subtask, which has 21 output classes, although findings hold more generally. Samples are chosen randomly. One sample per class is (naturally) necessary, and 2-5 samples per class will improve accuracy; accuracy stabilizes and will not improve significantly with more data.

% Then we perform comparison between different pruners from OBC family (see Figure \ref{fig:obc_comparison}). 
% SparseGPT being an approximation to exact OBC algorithm shows more significant drop in accuracy
% for the same sparsity level, whereas HybridOBC running the same procedure as exact OBC for all of the layers 
% but the one with the largest value of $d_{col}$ 
% (\texttt{fc1} layer for the case of vision transformer model) 
% yields roughly the same performance. These methods differ significantly in runtime
% - a single pruning step on \texttt{deit3\_base\_patch16\_224} takes 4 minutes for SparseGPT, 21
% minute for HybridOBC and 90 minutes for OBC. 
%% add timings

\subsection{Task Complexity and Model Compressibility}

\begin{figure}[!h]
    \begin{subfigure}{0.49\linewidth}
        \centering
        \includegraphics[width=\linewidth]{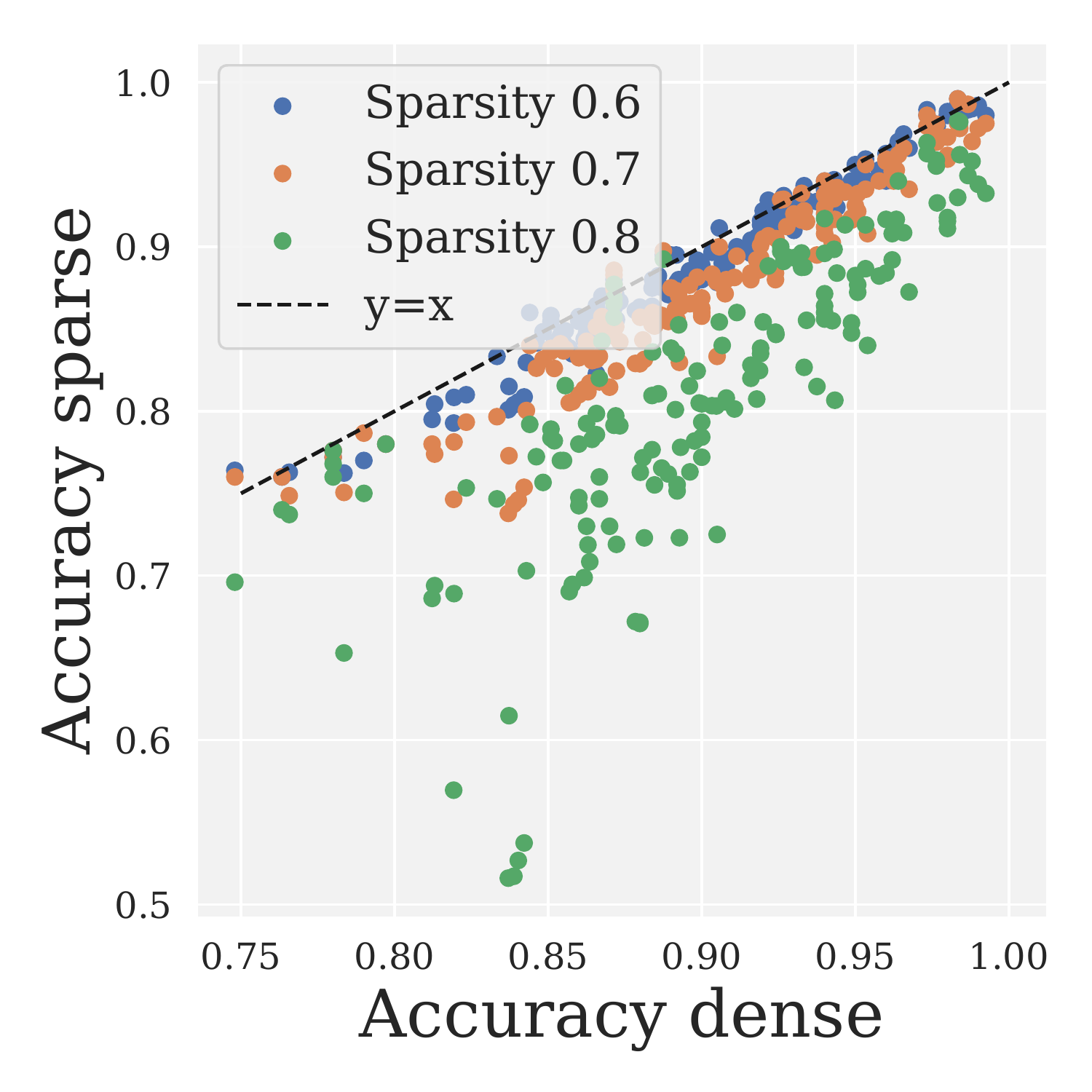}
    \end{subfigure}
    \begin{subfigure}{0.49\linewidth}
        \centering
        \includegraphics[width=\linewidth]{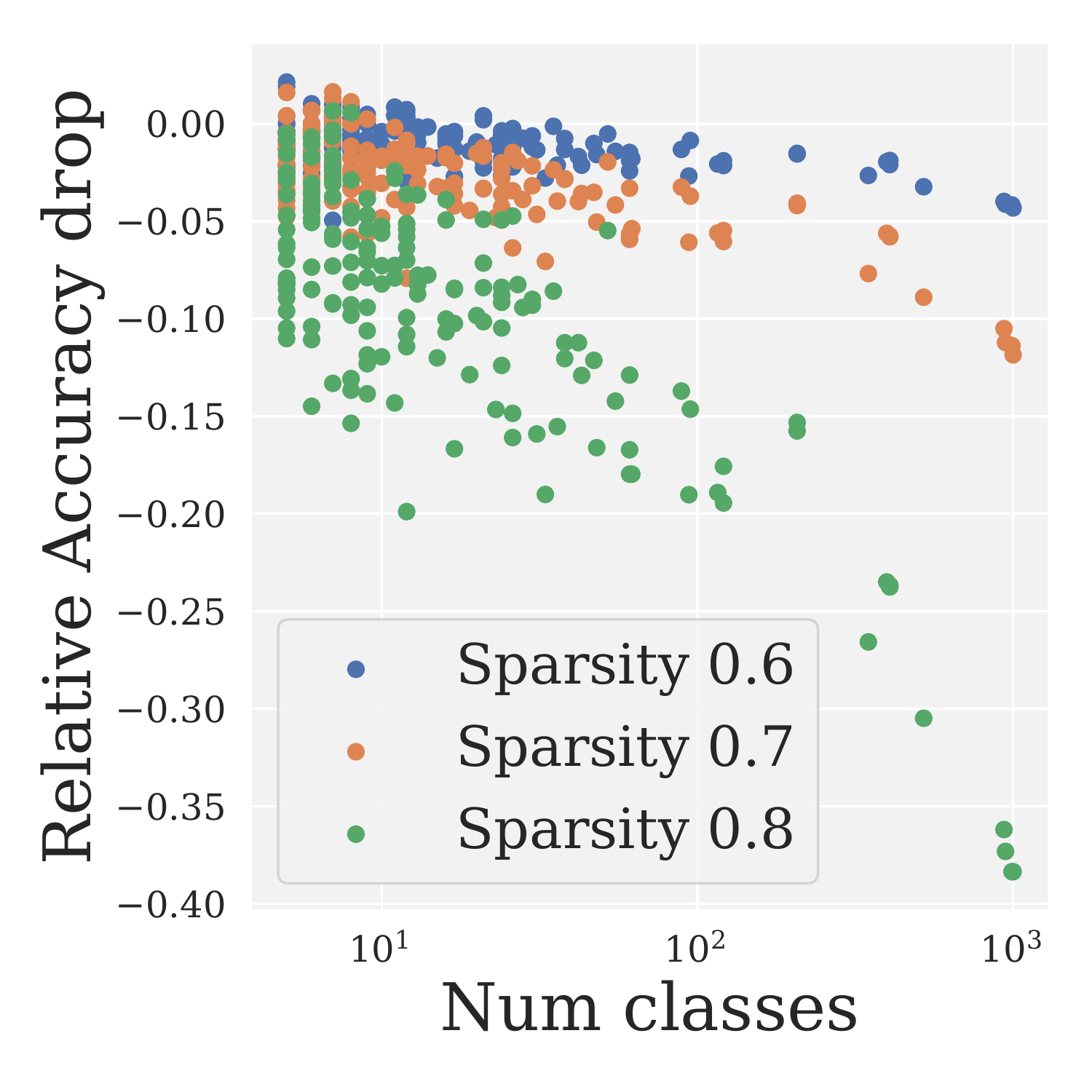}
    \end{subfigure}
    \caption{
        (\textbf{Left}) Accuracy on the ImageNet subtask for the dense model vs. accuracy of the sparse model.
        (\textbf{Right}) Relative accuracy drop vs. the number of classes.
    }
    \label{fig:compression_hierarchy}
    \vspace{-1em}
\end{figure}

\paragraph{Task Complexity.} In the next experiment, we run one-shot pruning over ImageNet with respect to the WordNet 
hierarchy~\cite{fellbaum2010wordnet}. We keep only the tasks containing at least 5 classes.
We selected three sparsity levels ($0.6, 0.7$, and $0.8$) and obtained single-step pruned models using TACO. 
(We use the FastOBC solver in this experiment, given the large number--166--of subsets.) 
The results of this large-scale experiment are provided in Figure~\ref{fig:compression_hierarchy}, 
where we examine the correlation between the accuracy of the sparse models relative to the corresponding dense one, as well as the relative accuracy drop post-compression, relative to the number of classes in the target specialization dataset. 

First, in the left-hand plot, we observe an apparent correlation between the accuracy of the uncompressed model and that of the sparse one on the given task. 
For most tasks, the performance drop is almost negligible for 60\%-70\% sparsity and becomes pronounced at 80\% sparsity only. 
Second, it is clear that certain subtasks allow for higher compression relative to others. This correlates with the number of target classes, which clearly measures task ``complexity.''
Specifically, the accuracy drop at a given sparsity appears to behave logarithmically with the number of classes. 
Finally, we note that for some subsets, the accuracy of the compressed model can be slightly superior to that of the original model. 

In addition, for each subtask selected, we evaluate the PTC model and compare the performance 
of PTC and TACO. Due to space constraints, results are presented in the Appendix; overall, they show that TACO outperforms
PTC on the majority of the tasks, and that the accuracy difference becomes more pronounced with an increase in sparsity. 
The difference 
is also more significant for smaller subtasks, showing the advantage of task specialization. 

% \begin{figure}[!h]
%     \begin{subfigure}{0.49\linewidth}
%         \centering
%         \includegraphics[width=\linewidth]{figures/compression_hierarchy/hier_pruning_ptc_vs_taco_all_sparsities_scatter.pdf}
%     \end{subfigure}
%     \begin{subfigure}{0.49\linewidth}
%         \centering
%         \includegraphics[width=\linewidth]{figures/compression_hierarchy/hier_pruning_taco_ptc_diff_all_sparsities_scatter.pdf}
%     \end{subfigure}
%     \caption{
%        (\textbf{Left}) Accuracy of PTC compressed model vs TACO on the specialized task (ImageNet subset).
%        (\textbf{Right}) Accuracy difference between models compressed by TACO and PTC on the specialized tasks.
%     }
%     \label{fig:compression_hierarchy_taco_vs_ptc}
%     \vspace{-1em}
% \end{figure}

\begin{figure}[!t]
    \begin{subfigure}{\linewidth}
        \centering
        \includegraphics[width=0.8\linewidth]{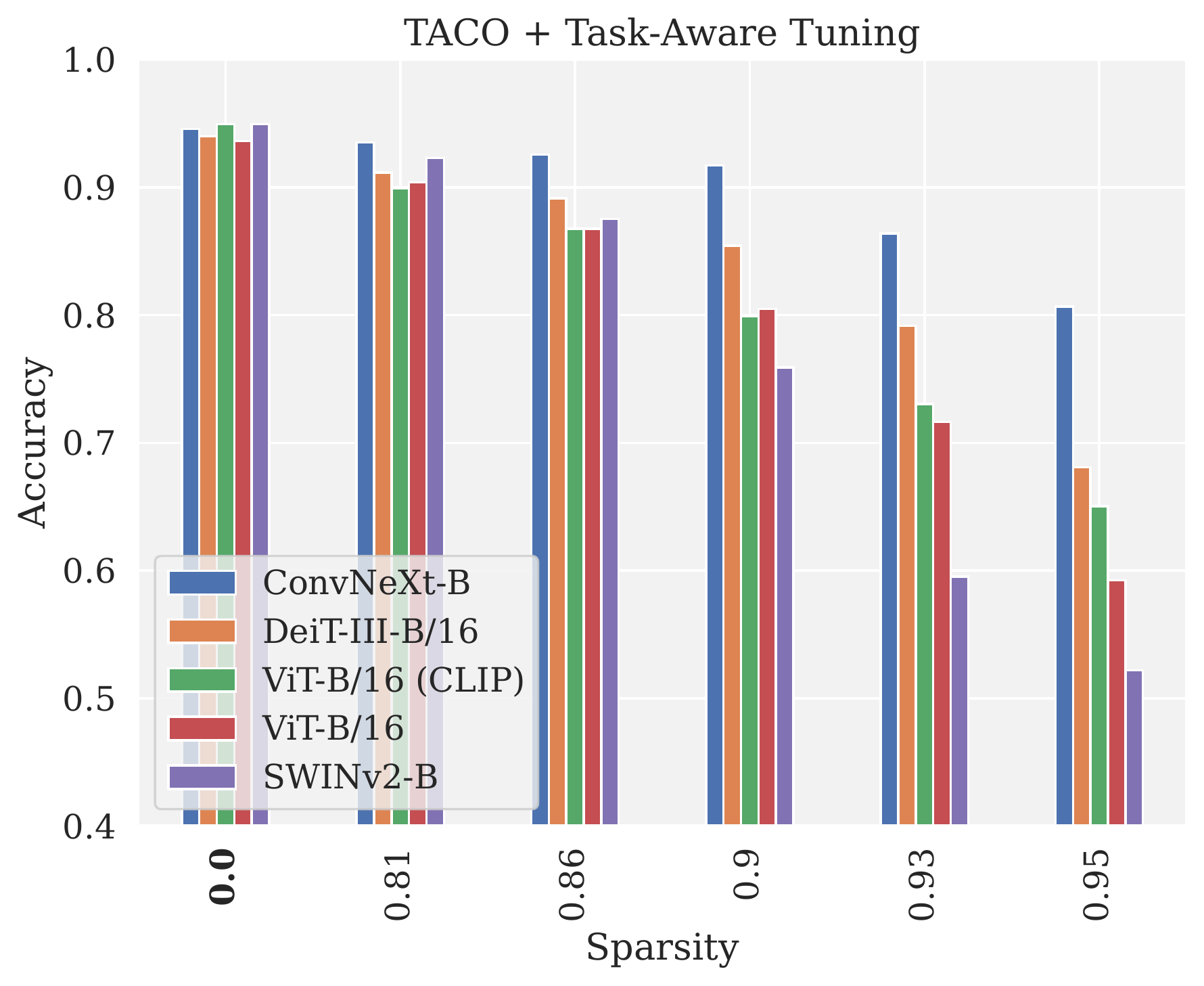}
    \end{subfigure}
    \caption{
        Accuracy of some SOTA models on the vehicle task. Note that all models have approximately the same number of parameters ($\approx86$ M parameters \emph{dense}). The first column  (0\% sparsity) is the dense model accuracy.
    }
    \label{fig:tuning_models}
    \vspace{-2em}
\end{figure}

\paragraph{The Impact of the Model Architecture.} 
Finally, we consider the impact of the model architecture on compressibility, 
keeping the target task constant. 
Specifically, we consider Transformer-based models (ViT/DeiT and SWIN) as well as the modern ConvNeXt architecture.
We instantiate these models in their ``base'' configuration, meaning that all have approximately the same number of dense parameters ( $\simeq$ 86M). 
Figure~\ref{fig:tuning_models} shows the results for the ``Vehicle'' subtask, with TACO + tuning. Results for other subtasks have  similar trends, and are deferred to the Appendix.

The first observation is that TACO is able to compress all models to $\sim$ 80\% sparsity without major accuracy loss. 
This is notable, since even the best known methods for ViT compression only achieve $\sim$ 50\% sparsity for general compression, before significant accuracy loss occurs~\cite{SVIT}. 
However, at higher sparsities there is a clear performance degradation for ViT-family models. 
The remarkable exception is the ConvNeXt model, which can reach 90\% sparsity with relatively low accuracy loss. 
One explanation is the model structure: we do not prune the depth-wise convolutions in ConvNeXt, as they have low computational and parameter count, this combination of inductive bias and domain knowledge appears to give this model a significant boost in terms of compressibility. 

\section{Extensions}

\subsection{Task-Aware Compression on iNaturalist}

To validate our approach in a different setup, 
we consider task-aware compression for the iNaturalist dataset~\cite{van2018inaturalist}. 
As a base model, we consider the ViT-Base model using the Masked AutoEncoder (MAE) pretraining setup~\cite{he2022masked}, 
which leads to state-of-the-art results for this general task.  
The only difference from MAE is that we finetune on iNaturalist21 rather than iNaturalist17. 

\begin{figure}[!h]
    \begin{subfigure}{\linewidth}
        \centering
        \includegraphics[width=\linewidth]{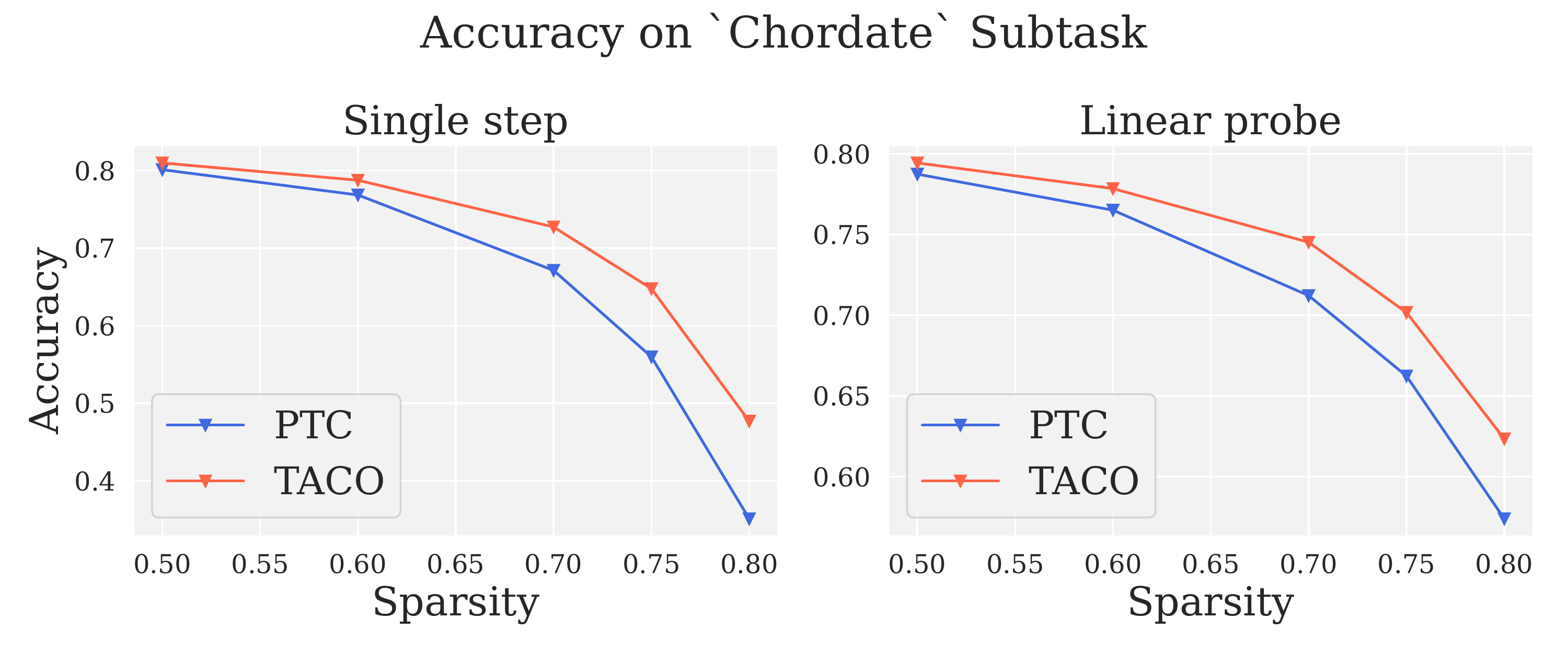}
    \end{subfigure}
    \caption{
        Unstructured pruning of the DeiT-III (B/16) model on iNaturalist subtasks.
        (\textbf{Left}) Accuracy before finetuning. 
        (\textbf{Right}) Accuracy after linear probing.
    }
    \label{fig:inaturalist_pruning}
    \vspace{-1em}
\end{figure}

%% Subset description would be better defered to the appendix part
The iNaturalist dataset has subclasses of different granularity, covering different ranges
of species. 
As individual tasks, we take subsets of the taxonomic tree from the following 
4 dataset categories: \texttt{kingdom}, \texttt{phylum}, \texttt{order}, \texttt{genus}. Then, we compared
the performance of task-agnostic and task-specific pruning. 
We use the HybridOBC solver, as it offers the best time-accuracy trade-off.
The results in Figure~\ref{fig:inaturalist_pruning}  show identical trends to ImageNet specialization: 
task-specific pruning again performs significantly better compared to task-agnostic pruning.
We provide  additional experiments, for other taxonomic levels and subgroups, in the Appendix. 
% Furthermore, the more specific the category (the Contruroides genus is more narrow than the Chordate phylum), the more pronounced the relative accuracy drop between dense and sparse models. 

\subsection{TACO for Pruning and Quantization}

Our approach can be 
combined with other means of compression such as quantization to lower precision 
in order to achieve higher speed-ups. 

\begin{figure}[htb]
    \centering
    \begin{subfigure}{0.49\linewidth}  
        \includegraphics[width=\linewidth]{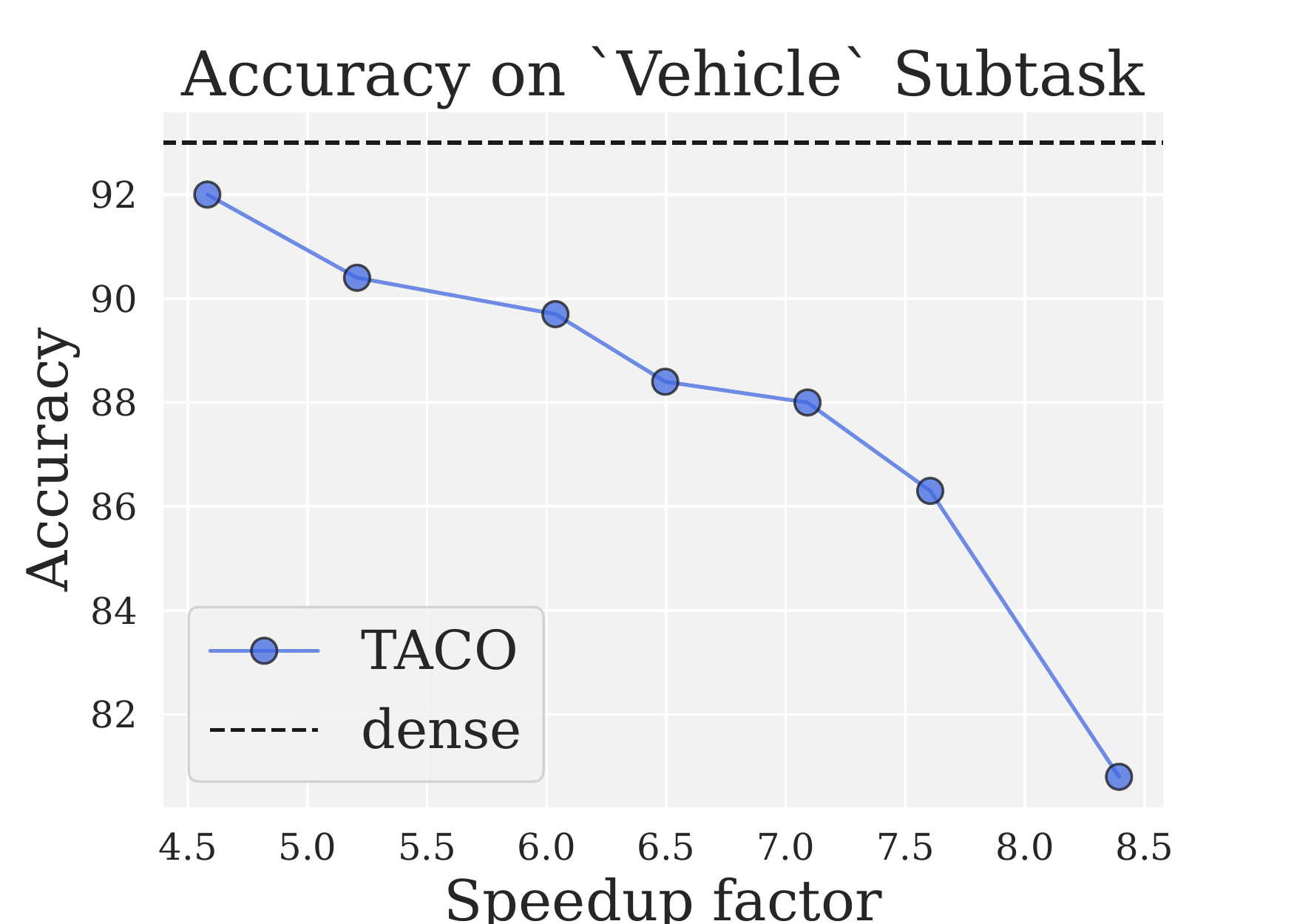}
    \end{subfigure}
    \begin{subfigure}{0.49\linewidth}  
        \includegraphics[width=\linewidth]{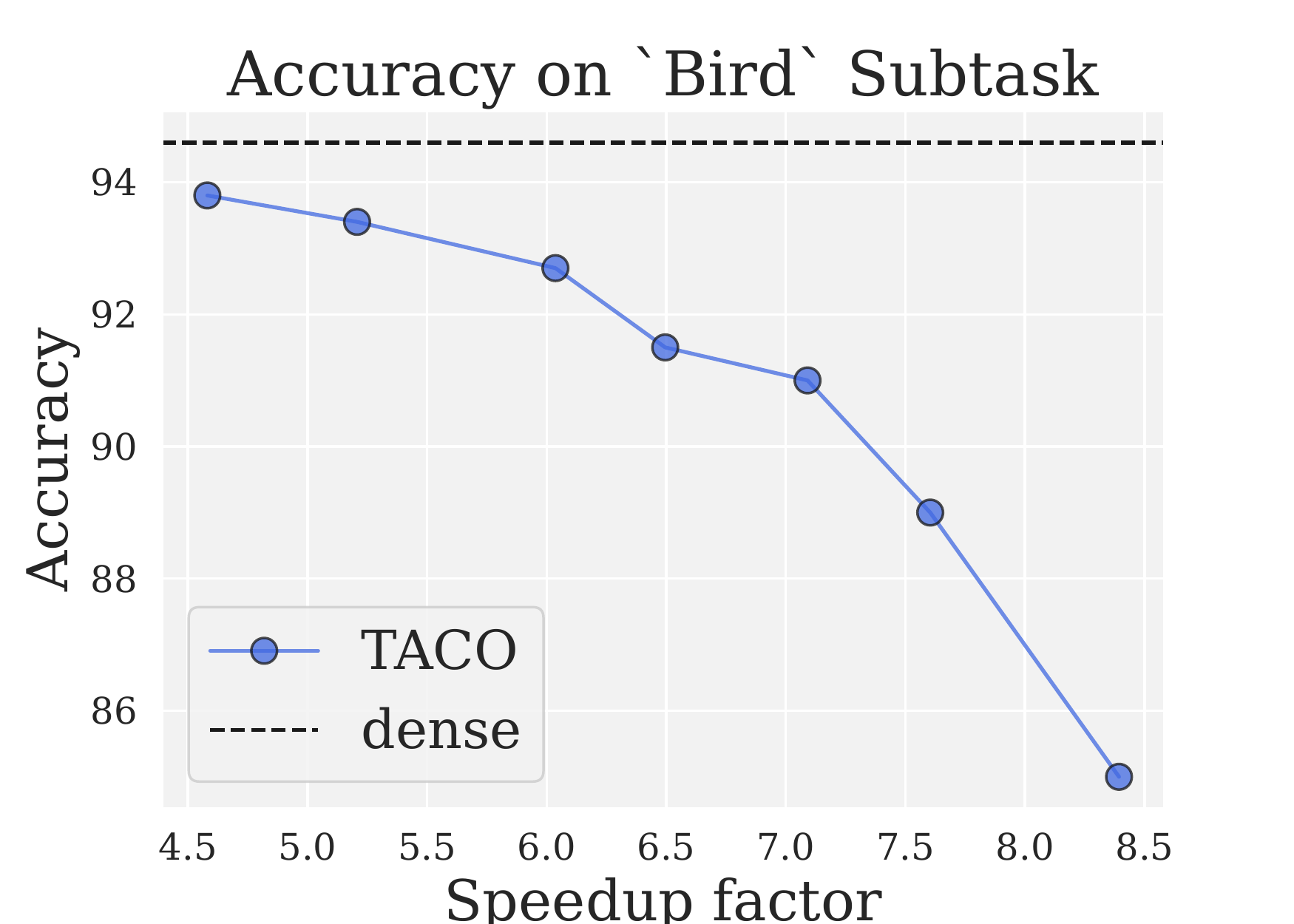}
    \end{subfigure}
    \caption{
        CPU Speedup on ResNet50 vs accuracy for block4 compression with 8-bit quantization, on the DeepSparse runtime.
        The dashed line shows the performance of the original dense model.
    }
    \label{fig:qat_speedups}
\end{figure}

To illustrate this, we consider a pretrained ResNet50/ImageNet-1K model, and prune it with TACO using the OBC solver (since it is more accurate and the model is relatively small)
using \texttt{block4} sparsity, where each group of 4 weights is either zero or nonzero. We then quantize the
model to 8-bit precision via the GPTQ quantizer~\cite{frantar2023optq}; finally, we finetune the sparse quantized model with Quantization-Aware Training (QAT). The details
of the QAT finetuning procedure are provided in Appendix \ref{app:finetuning_details}. 
The speedup-accuracy trade-offs are given in Figure \ref{fig:qat_speedups}.

Specifically, compounding block-sparsity and quantization allows us to achieve 4-5x speed-up at relatively negligible loss of performance, 
and up to 8-9x speed-up at moderate accuracy degradation.

\paragraph{Higher Compresion for ``Easier'' Tasks.} For relatively simpler subtasks, with lower number of target classes, this approach works especially well. 
We illustrate this in Figure~\ref{fig:qat_speedups_easy} below.  

\begin{figure}[htb]
    \centering
    \begin{subfigure}{0.49\linewidth}  
        \includegraphics[width=\linewidth]{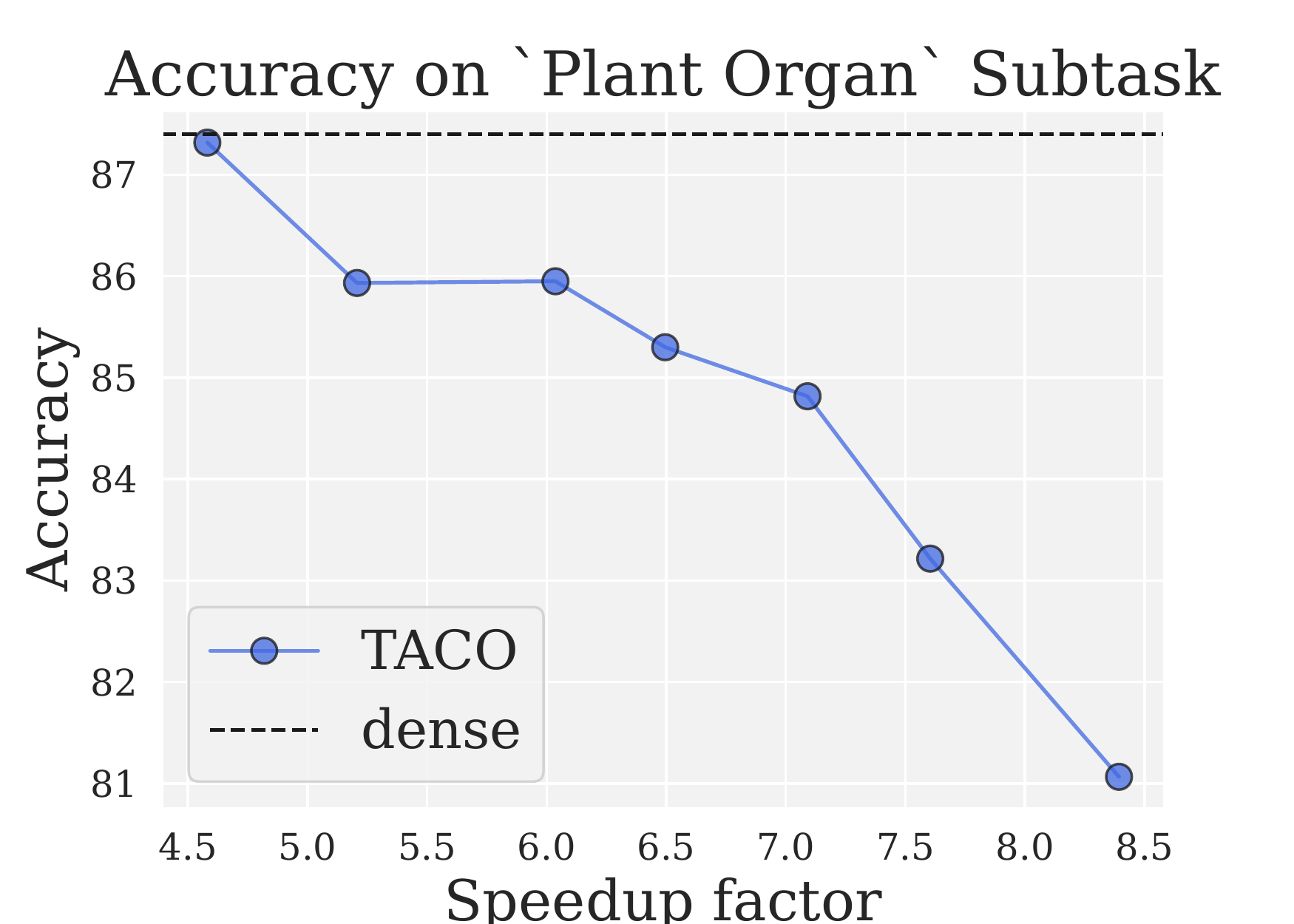}
    \end{subfigure}
    \begin{subfigure}{0.49\linewidth}  
        \includegraphics[width=\linewidth]{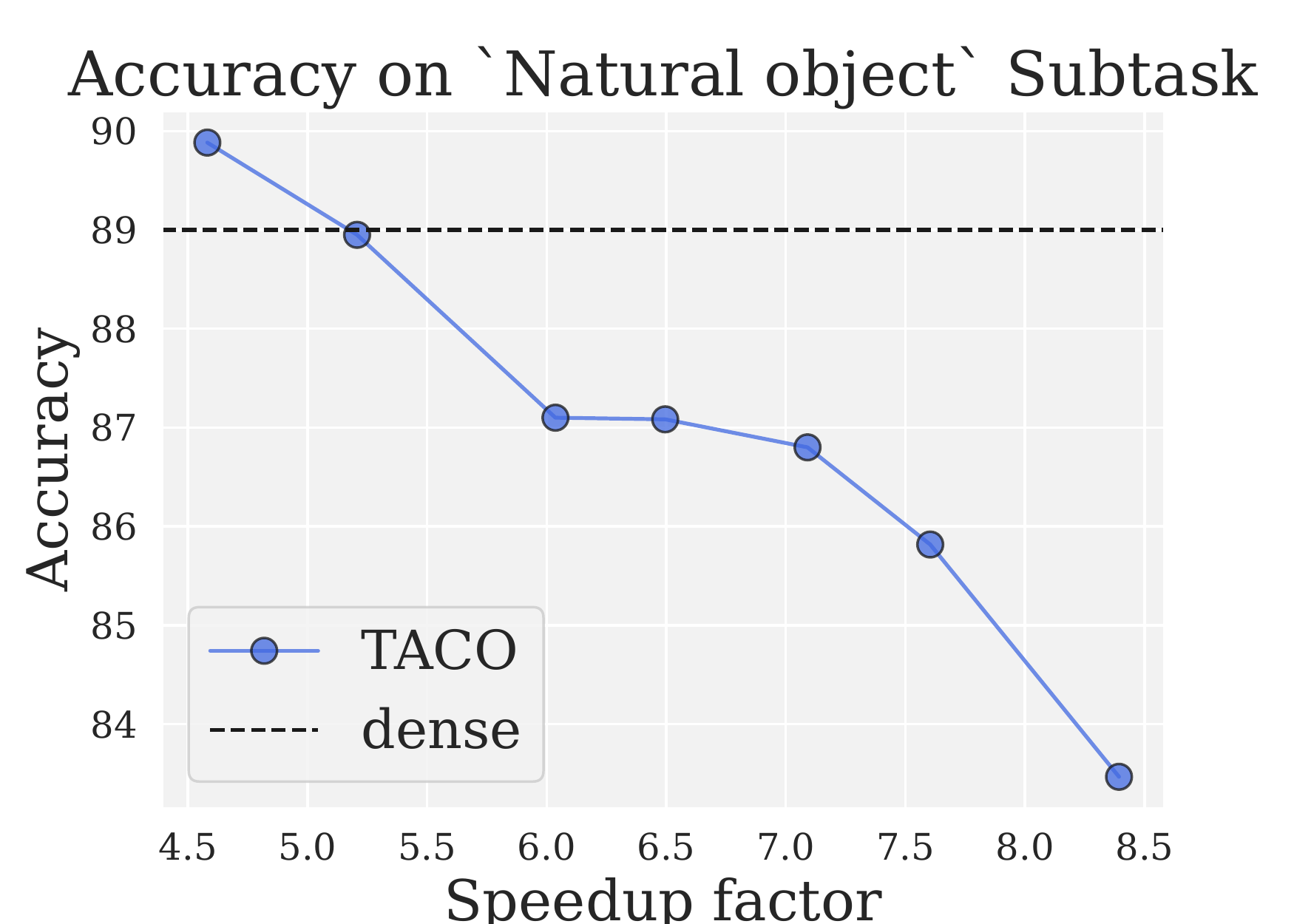}
    \end{subfigure}
    \caption{
        CPU Speedup on ResNet50 vs accuracy for block4 compression with 8-bit quantization.
        Dashed line depicts the performance of the dense model.
    }
    \label{fig:qat_speedups_easy}
\end{figure}

Specifically, one can achieve 4.5-5x speed-up without \emph{any
performance loss}, compared to the original model, and still have good accuracy ($<$ 10\% relative accuracy drop) at close to  compression rates.

\subsection{Task-Aware Structured Pruning}

\begin{figure}[!h]
    \begin{subfigure}{\linewidth}
        \centering
        \includegraphics[width=\linewidth]{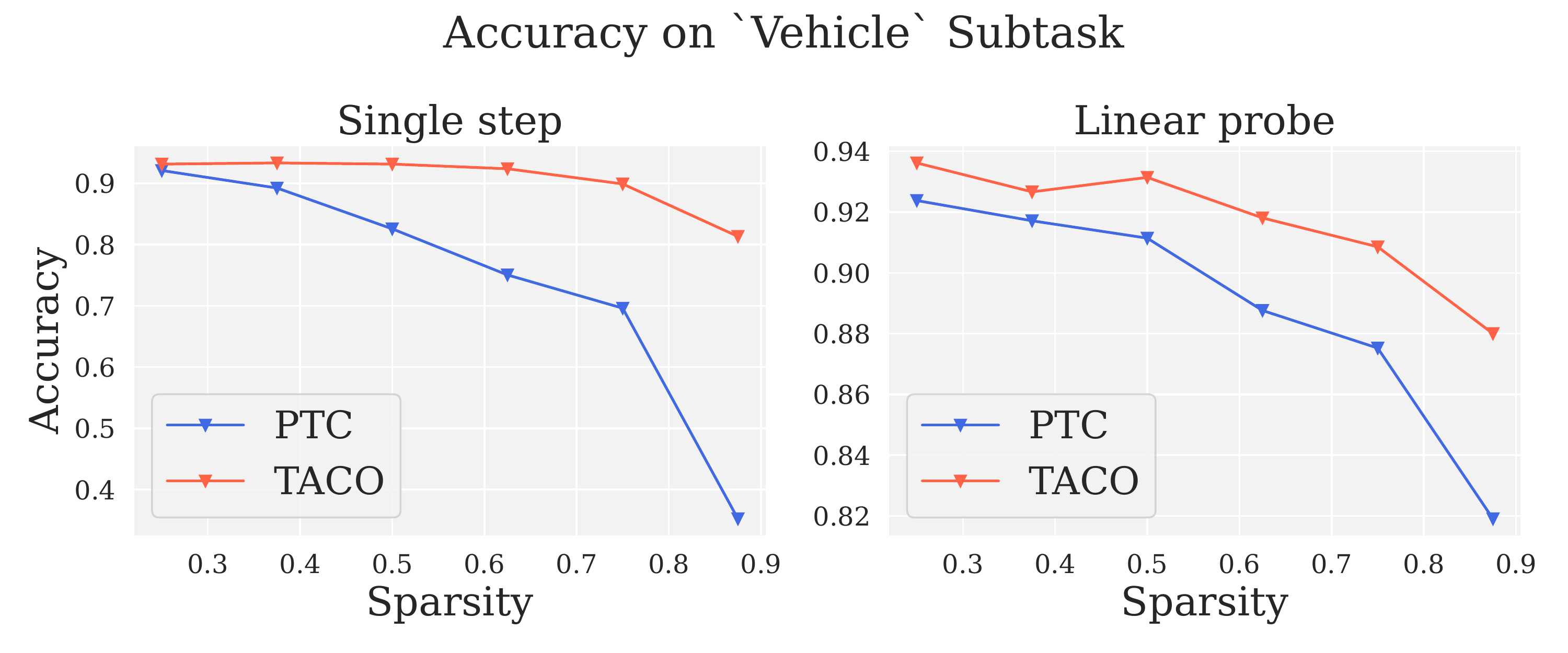}
    \end{subfigure}
    \caption{ 
         Structured pruning of DeiT-III (B/16).
        (\textbf{Left}) Single-step accuracy. 
        (\textbf{Right}) Accuracy after finetuning.
    }
    \label{fig:structured_pruning_ft}
    \vspace{-1em}
\end{figure}

The prior experiments performed fine-grained, unstructured pruning. 
We now investigate our approach in the context of \emph{structured, feature-level} pruning: specifically, we return to the ImageNet setup, and use a task-specific calibration set, the ZipLM structured pruning solver~\cite{kurtic2023ziplm} applied uniformly to the hidden dimension in the fully-connected layers, and the DeiT-III-B/16(224)
model. The results in Figure~\ref{fig:structured_pruning_ft} show the same trends as for unstructured pruning: with finetuning, we can achieve up to 50\% structured sparsity 
 (in FC-layers) with moderate accuracy loss, doubling compression relative to general post-training pruning.

\begin{figure}[!h]
    \centering
    \begin{subfigure}{\linewidth}
        \includegraphics[width=\linewidth]{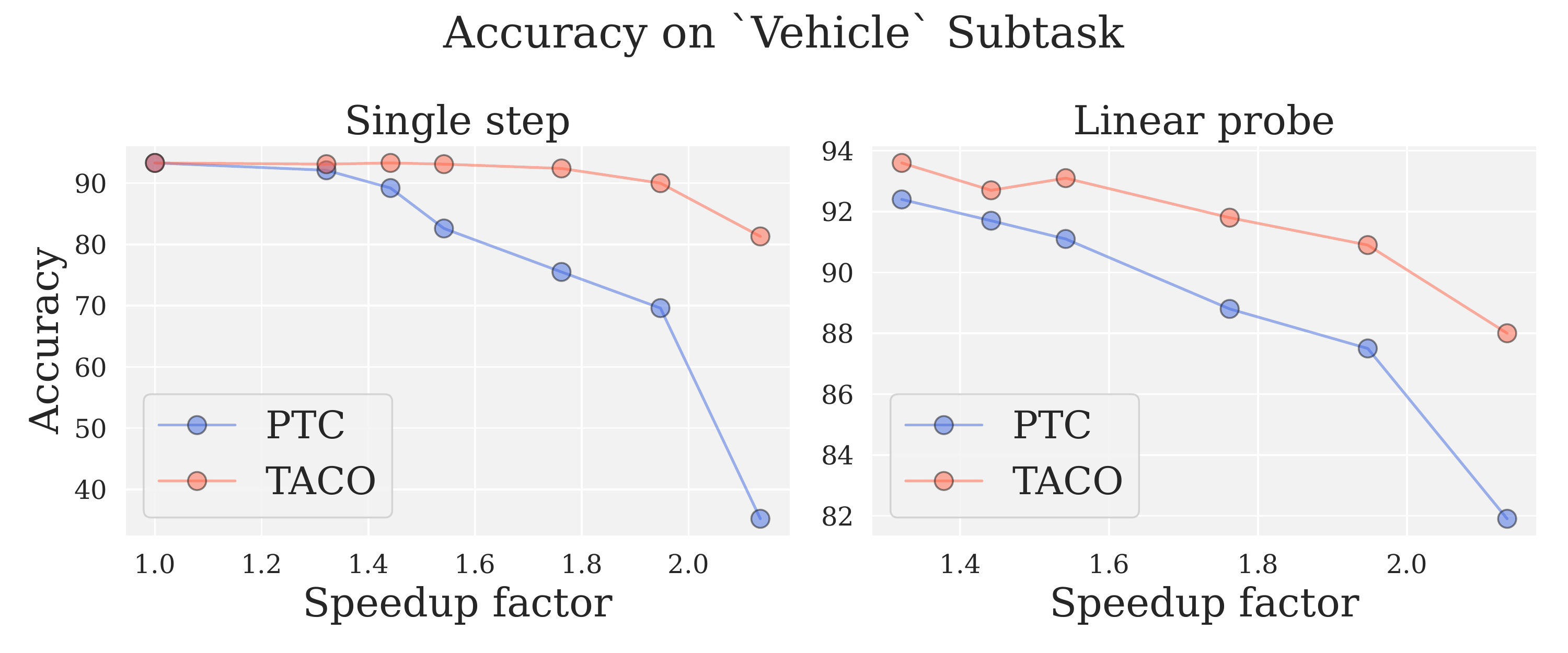}
    \end{subfigure}
    \caption{GPU Speedup on DeiT-III (B/16) vs. accuracy for structured compression.}
    \label{fig:gpu_speedup}
    \vspace{-1em}
\end{figure}

\paragraph{GPU Inference Speedups.} We now show that this setup leads to significant speedups on GPU inference. 
Specifically, we executed the structured-pruned models obtained above in Pytorch, on an NVIDIA T4 GPU,
and measure the speedups, relative to accuracy when specializing on specific subtasks, shown in Figure~\ref{fig:gpu_speedup}. 
(The speedups are almost identical on any subtask, as the classification head has a negligible cost.) 
We note that we can obtain a 50\% inference speedup with negligible accuracy loss, and 2x speedup while maintaining accuracy $\geq 90\%$ on this particular task and model combination.

\subsection{Transfer Learning Performance \\ and  Comparison with Lottery Tickets}
\label{sec:transfer}
% \vspace{-1em}

\begin{algorithm}[H]
    \small
    \caption{Overview of gradual transfer learning setup.}
    \begin{algorithmic}[1]
    \State \textbf{Input}: Dense model $M$, Sparsity target $\sigma$, Pruning method $\Gamma$.
    % \State Set the initial model $M$ to be dense.
    \Repeat
    \State Prune $50\%$ of remaining weights in $M$ via method $\Gamma$. 
    \State Update $M$ to the pruned model.
    \State Finetune $M$ for $25$ epochs, maintaining the sparsity mask.

    \Until{$M$'s sparsity reaches some target level $\sigma$.}
    \State Return $M$.
    \end{algorithmic}
    \label{alg:gradualTACO}
\end{algorithm}
\vspace{-0.5em}

We now examine the performance of TACO in the more challenging \emph{transfer learning} setting, 
in which the target specialized task is no longer a subset of the pretraining dataset.
Here, we would expect the pretrained model to have non-random performance on the target task, 
but its accuracy should be improved via fine-tuning. 
Due to this difference, we consider two applications of TACO, using a sample of transfer task data as the calibration set: 
we can apply TACO \emph{initially}, in a single step, followed by finetuning, or \emph{gradually}, in steps of increasing sparsity, separated by finetuning.  
The gradual approach (Algorithm~\ref{alg:gradualTACO}) coincides with Lottery Tickets for transfer learning~\cite{chen2021lottery}, and thus provides a good point of comparison relative to this method.

\begin{figure}[!h]
\label{fig:lth-comparison}
\subfloat{\includegraphics[width=0.52\linewidth]{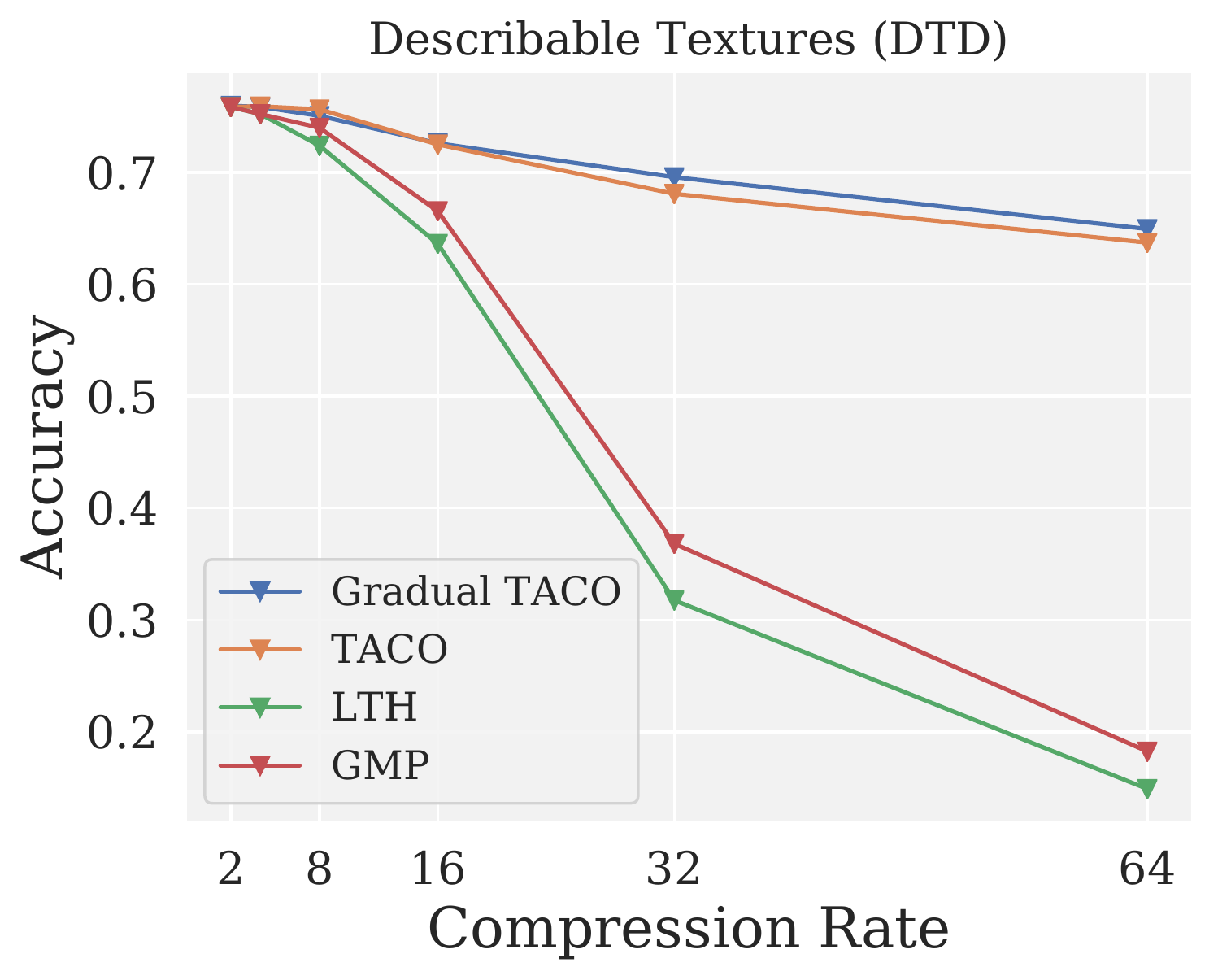}}
\subfloat{\includegraphics[width=0.5\linewidth]{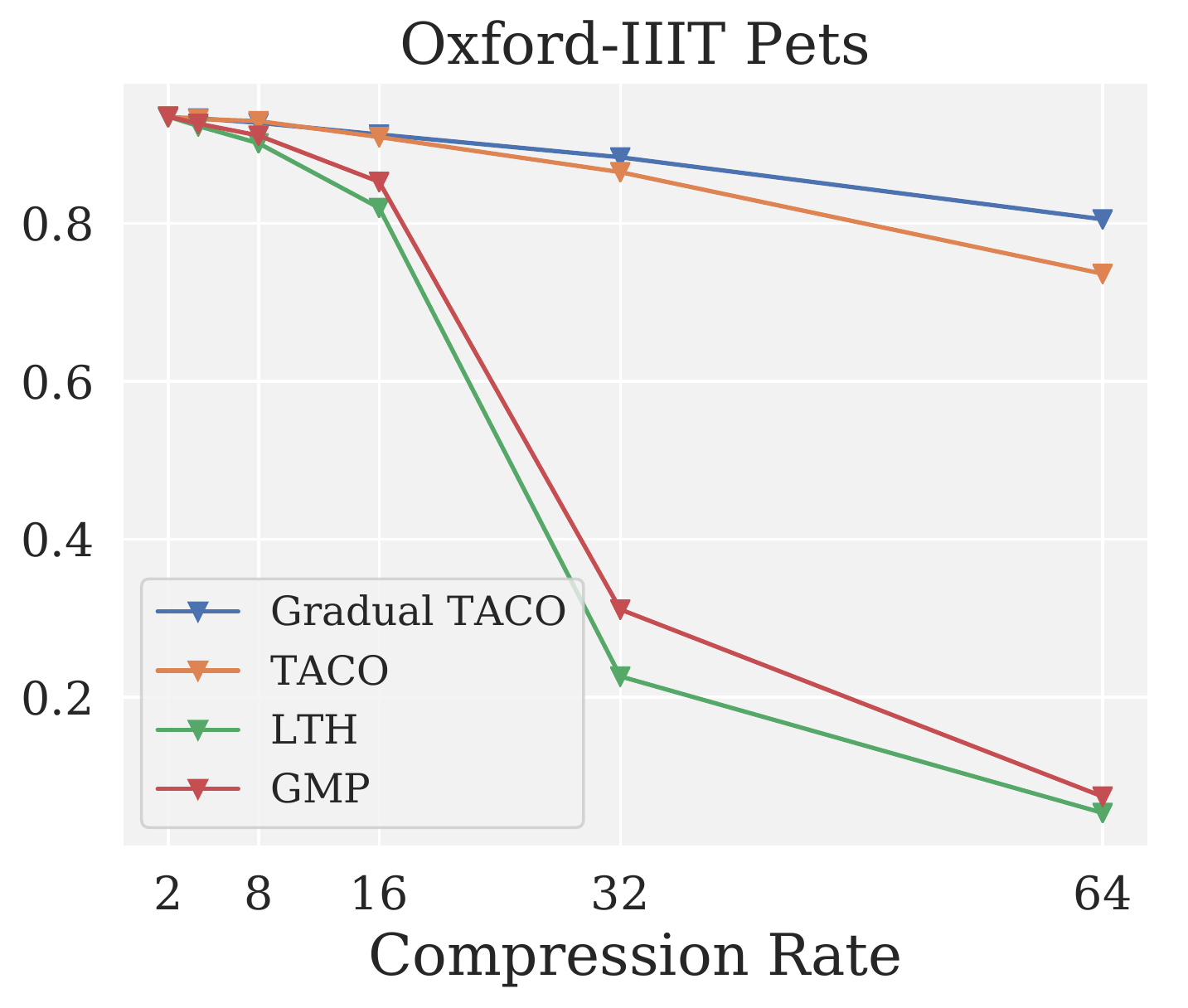}}

\subfloat{\includegraphics[width=0.52\linewidth]{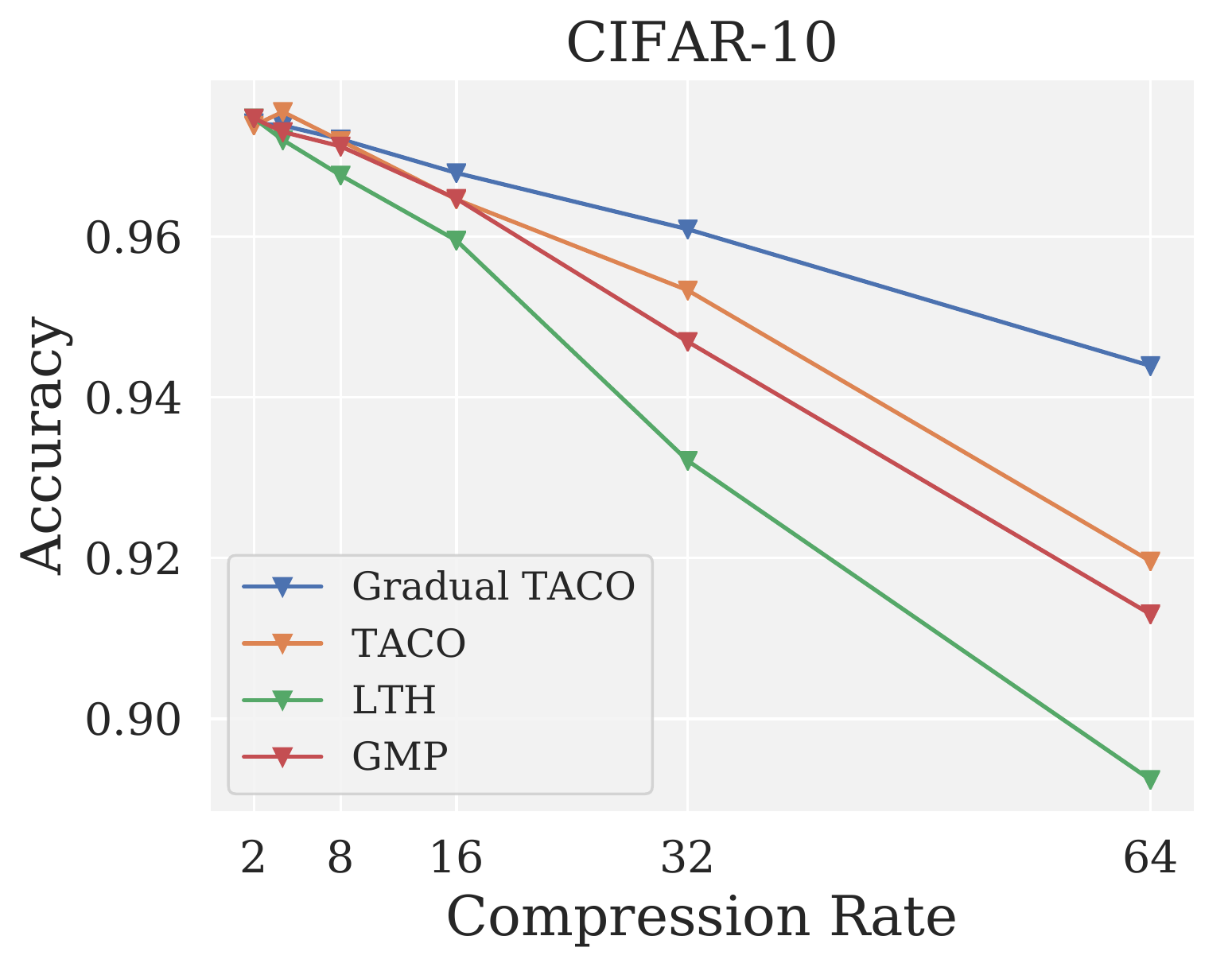}}
\subfloat{\includegraphics[width=0.5\linewidth]{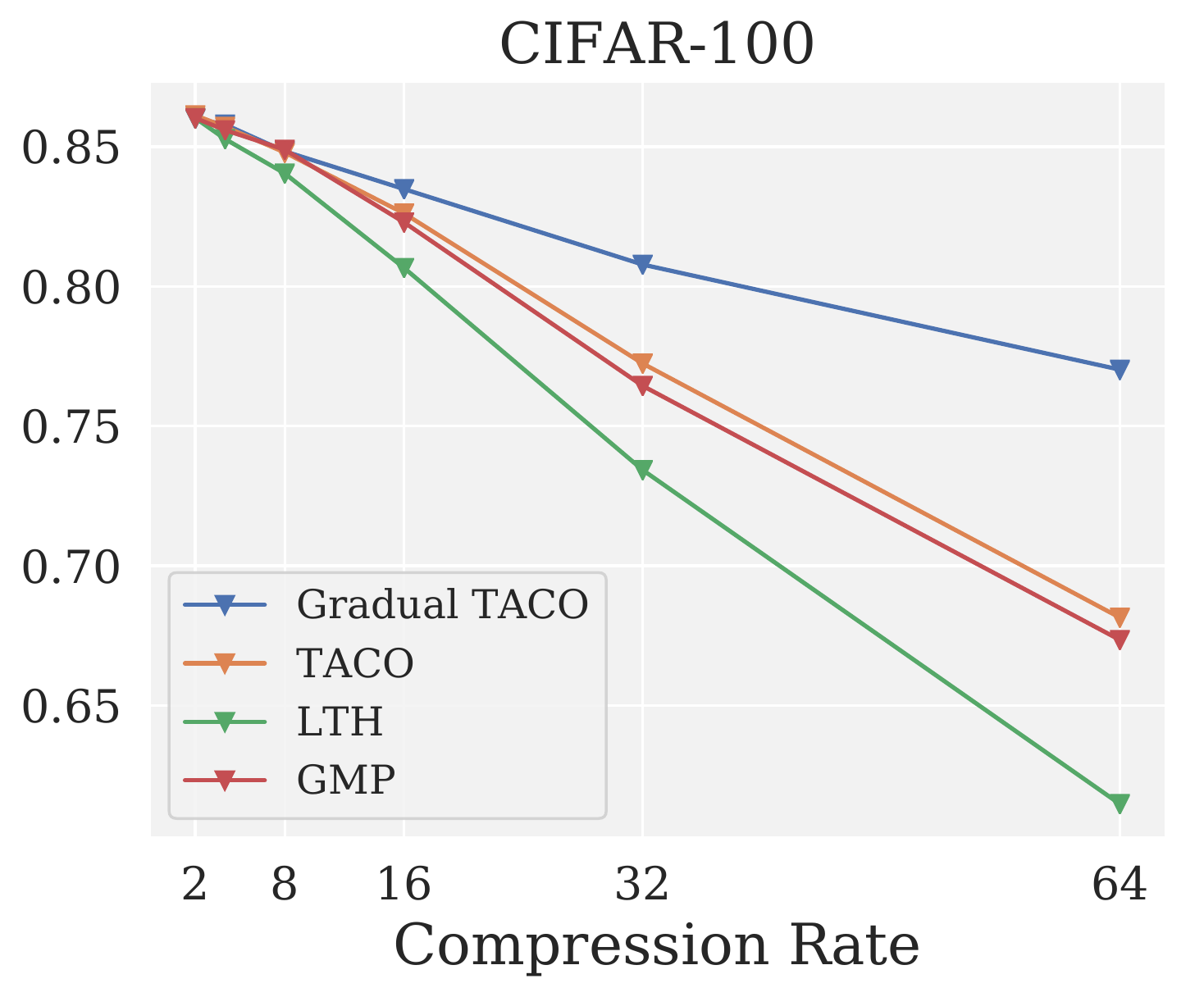}}

\subfloat{\includegraphics[width=0.52\linewidth]{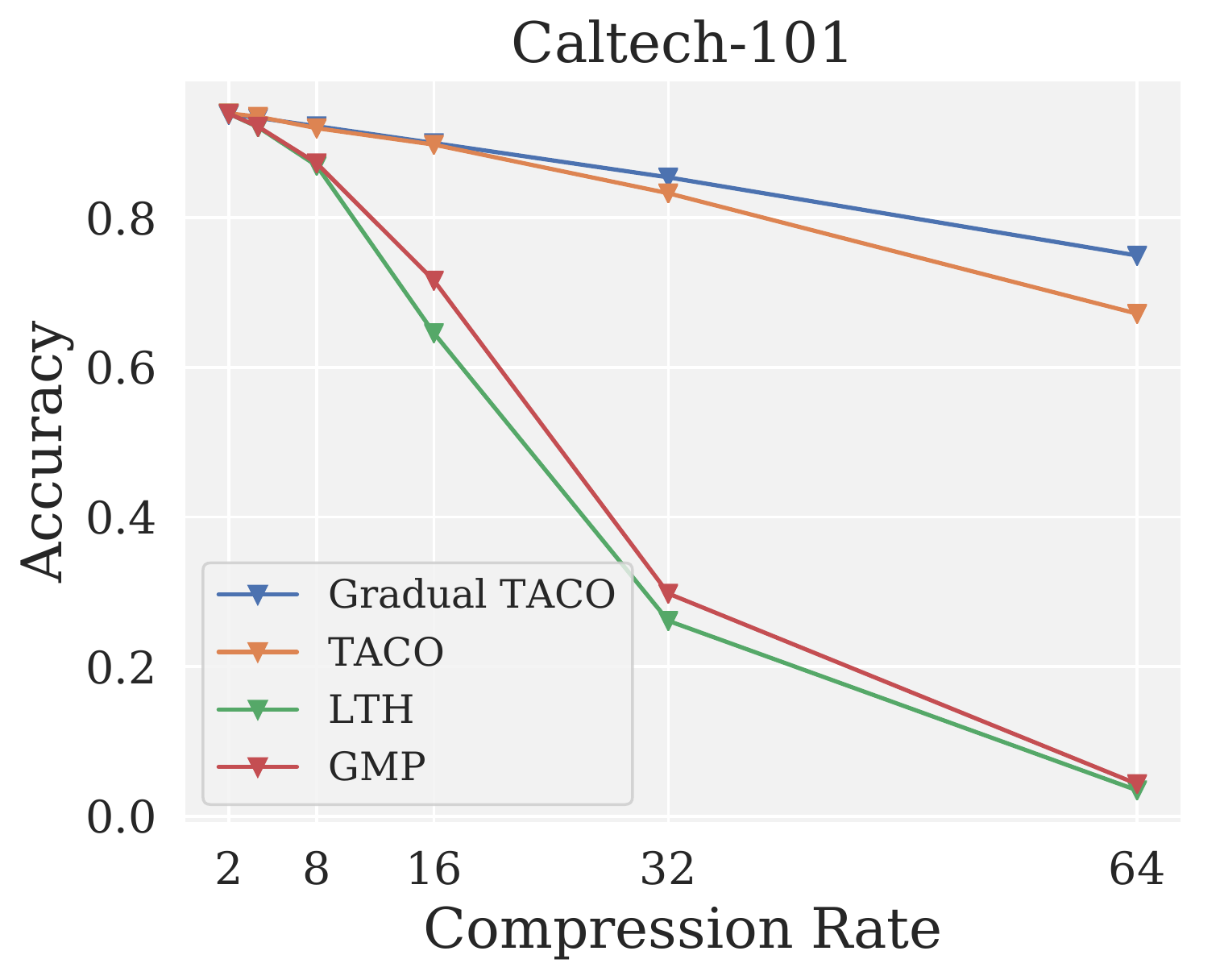}}   
\subfloat{\includegraphics[width=0.5\linewidth]{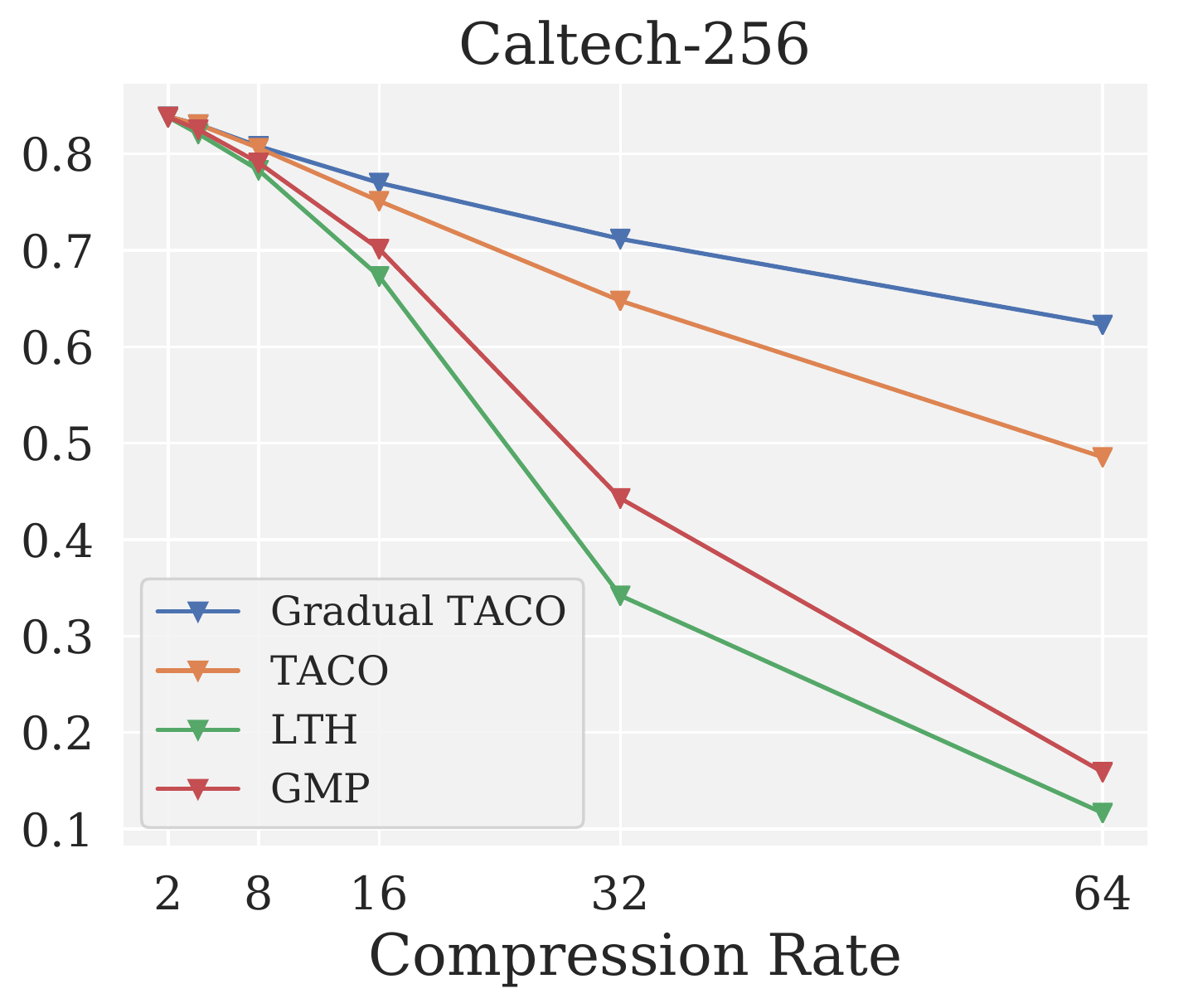}}
\caption{Experiments for model compression in a standard transfer learning setup. We compare TACO and Gradual TACO with Lottery Tickets for Transfer (LTH) and Gradual Magnitude Pruning (GMP) in terms of accuracy (Y axis) versus model size reduction (X-axis).}\label{fig:transfer}
\end{figure}

Following~\cite{chen2021lottery}, we use a pretrained ResNet50/ImageNet-1K model as the base, and run TACO on a subset of 1024 samples randomly selected from the transfer task. 
In the gradual case, each pruning step removes 50\% of the remaining model weights, followed by 25 epochs of finetuning on the transfer task. 
In the single-step case, we prune once, and fine-tune for the cumulative equivalent number of epochs.
We use exactly the same setting with gradual magnitude pruning (GMP) and Lottery Tickets for Transfer (LTH)~\cite{chen2021lottery} methods, applying the corresponding pruner at the compression step.  
Figure \ref{fig:transfer} shows results for 6 standard transfer learning downstream datasets~\cite{kornblith2019better}, for compression rates between 2 (50\% sparsity) and 64 (98.5\% sparsity). 

Across all tasks and sparsities, Gradual TACO achieves highest accuracy, followed by single-step TACO + finetuning. 
Remarkably, Gradual TACO can achieve moderate accuracy loss even for $\geq$ 95\% sparsity, 
outperforming the GMP and LTH data-agnostic methods by a very large margin. 
The strong performance of Single-step TACO shows that model specialization can occur even when there is just a single compression step, 
and the calibration data is \emph{not} part of the pretraining task. 
This suggests that model specialization may also happen for tasks that are not strictly in-distribution, which bodes well for the generality of the method. 
At the same time, the major accuracy difference between TACO and LTH/GMP suggests that task-aware specialization provides significantly better compression, relative to techniques based on magnitude pruning and fine-tuning, which are task-agnostic. 

\vspace{-0.3em}
\section{Related Work}
\vspace{-0.3em}

The area of model compression has seen significant activity recently, given the rapid rise in computational and memory costs for modern deep models~\cite{hoefler2021sparsity, gholami2021survey}. 

Our work is closest to the line of research on \emph{post-training compression}, with methods for single-step pruning~\cite{hubara2021accelerated, frantar-obc, sparseGPT} or quantization, e.g.~\cite{nagel2020up, li2021brecq, frantar2023optq}. 
As we have shown, solvers proposed by this line of work can be directly integrated into the TACO framework. 
The key novelty in our work is the idea of \emph{model specialization} on narrower tasks, which we have shown to occur consistently for both in-distribution and (moderately) out-of-distribution tasks. In turn, this allows our method to achieve much higher compression ratios relative to standard, general post-training compression, at the same level of accuracy. 
Another source of novelty relative to this line of work is the fact that we show for the first time that the calibration set can be used successfully for finetuning, at minor computational cost, and without overfitting. 

Another related line of work is that on inducing sparsity in the context of transfer learning tasks~\cite{chen2020lottery, chen2021lottery, zafrir2021prune, iofinova2022well}. 
Some work in this area~\cite{zafrir2021prune, iofinova2022well} considers the orthogonal case in which models are pruned on the pretraining task; this would require significant additional computation. 
By contrast, we consider a setting in which compression should be performed extremely efficiently, and without additional hyper-parameters. 
The Lottery Ticket for Transfer (LTH) work aims for a similar goal but does so in a data-agnostic way. 
As we showed in Section~\ref{sec:transfer}, our approach can very clearly outperform LTH, especially at the larger compression levels required to obtain practical speedups. 
% \vspace{-0.5em}
\section{Conclusions}
% \vspace{-0.5em}

We have shown for the first time that pretrained vision models can be efficiently specialized via task-aware compression. 
Our approach holds across pretraining datasets, target tasks, and model families, providing significantly higher compression, at the same accuracy level, 
relative to generic post-training compression, specifically because of model specialization during compression. 
Future work could investigate other compression types, including quantization, as well as out-of-distribution specialization.  

\section{Acknowledgements}

The authors would like to thank Eugenia Iofinova and Alexandra Peste for useful discussions at the inception of this project. The research was supported by the Scientific Service Units (SSU) 
of IST Austria through resources provided by Scientific Computing (SciComp).
D.K., who carried out experiments on TACO comparison in one-shot and linear finetune setting 
and measured hardware speedups, was supported by Russian Science Foundation, grant 21-11-00373.

% \newpage

\bibliography{main.bib}

\begin{thebibliography}{10}

\bibitem{beery2018recognition}
S.~Beery, G.~Van~Horn, and P.~Perona.
\newblock Recognition in terra incognita.
\newblock In {\em Proceedings of the European conference on computer vision
  (ECCV)}, pages 456--473, 2018.

\bibitem{DINO}
M.~Caron, H.~Touvron, I.~Misra, H.~J\'egou, J.~Mairal, P.~Bojanowski, and
  A.~Joulin.
\newblock Emerging properties in self-supervised vision transformers.
\newblock In {\em Proceedings of the IEEE/CVF International Conference on
  Computer Vision (ICCV)}, pages 9650--9660, October 2021.

\bibitem{SVIT}
T.~Chen, Y.~Cheng, Z.~Gan, L.~Yuan, L.~Zhang, and Z.~Wang.
\newblock Chasing sparsity in vision transformers: An end-to-end exploration.
\newblock {\em Advances in Neural Information Processing Systems},
  34:19974--19988, 2021.

\bibitem{chen2021lottery}
T.~Chen, J.~Frankle, S.~Chang, S.~Liu, Y.~Zhang, M.~Carbin, and Z.~Wang.
\newblock The lottery tickets hypothesis for supervised and self-supervised
  pre-training in computer vision models.
\newblock In {\em Proceedings of the IEEE/CVF Conference on Computer Vision and
  Pattern Recognition}, pages 16306--16316, 2021.

\bibitem{chen2020lottery}
T.~Chen, J.~Frankle, S.~Chang, S.~Liu, Y.~Zhang, Z.~Wang, and M.~Carbin.
\newblock The lottery ticket hypothesis for pre-trained bert networks.
\newblock {\em Advances in neural information processing systems},
  33:15834--15846, 2020.

\bibitem{chen2020simple}
T.~Chen, S.~Kornblith, M.~Norouzi, and G.~Hinton.
\newblock A simple framework for contrastive learning of visual
  representations.
\newblock In {\em International conference on machine learning}, pages
  1597--1607. PMLR, 2020.

\bibitem{ViT}
A.~Dosovitskiy, L.~Beyer, A.~Kolesnikov, D.~Weissenborn, X.~Zhai,
  T.~Unterthiner, M.~Dehghani, M.~Minderer, G.~Heigold, S.~Gelly, J.~Uszkoreit,
  and N.~Houlsby.
\newblock An image is worth 16x16 words: Transformers for image recognition at
  scale, 2020.

\bibitem{robustness}
L.~Engstrom, A.~Ilyas, H.~Salman, S.~Santurkar, and D.~Tsipras.
\newblock Robustness (python library), 2019.

\bibitem{evci2019difficulty}
U.~Evci, F.~Pedregosa, A.~Gomez, and E.~Elsen.
\newblock The difficulty of training sparse neural networks.
\newblock {\em arXiv preprint arXiv:1906.10732}, 2019.

\bibitem{fellbaum2010wordnet}
C.~Fellbaum.
\newblock Wordnet.
\newblock In {\em Theory and applications of ontology: computer applications},
  pages 231--243. Springer, 2010.

\bibitem{sparseGPT}
E.~Frantar and D.~Alistarh.
\newblock Sparsegpt: Massive language models can be accurately pruned in
  one-shot, 2023.

\bibitem{frantar2023optq}
E.~Frantar, S.~Ashkboos, T.~Hoefler, and D.~Alistarh.
\newblock {OPTQ}: Accurate quantization for generative pre-trained
  transformers.
\newblock In {\em The Eleventh International Conference on Learning
  Representations}, 2023.

\bibitem{frantar-obc}
E.~Frantar, S.~P. Singh, and D.~Alistarh.
\newblock Optimal brain compression: A framework for accurate post-training
  quantization and pruning.
\newblock In A.~H. Oh, A.~Agarwal, D.~Belgrave, and K.~Cho, editors, {\em
  Advances in Neural Information Processing Systems}, 2022.

\bibitem{gholami2021survey}
A.~Gholami, S.~Kim, Z.~Dong, Z.~Yao, M.~W. Mahoney, and K.~Keutzer.
\newblock A survey of quantization methods for efficient neural network
  inference.
\newblock {\em arXiv preprint arXiv:2103.13630}, 2021.

\bibitem{han2015deep}
S.~Han, H.~Mao, and W.~J. Dally.
\newblock Deep compression: Compressing deep neural networks with pruning,
  trained quantization and huffman coding.
\newblock {\em arXiv preprint arXiv:1510.00149}, 2015.

\bibitem{he2022masked}
K.~He, X.~Chen, S.~Xie, Y.~Li, P.~Doll{\'a}r, and R.~Girshick.
\newblock Masked autoencoders are scalable vision learners.
\newblock In {\em Proceedings of the IEEE/CVF Conference on Computer Vision and
  Pattern Recognition}, pages 16000--16009, 2022.

\bibitem{he2016deep}
K.~He, X.~Zhang, S.~Ren, and J.~Sun.
\newblock Deep residual learning for image recognition.
\newblock In {\em Proceedings of the IEEE conference on computer vision and
  pattern recognition}, pages 770--778, 2016.

\bibitem{hoefler2021sparsity}
T.~Hoefler, D.~Alistarh, T.~Ben-Nun, N.~Dryden, and A.~Peste.
\newblock Sparsity in deep learning: Pruning and growth for efficient inference
  and training in neural networks.
\newblock {\em Journal of Machine Learning Research}, 22(241):1--124, 2021.

\bibitem{hubara2021accelerated}
I.~Hubara, B.~Chmiel, M.~Island, R.~Banner, S.~Naor, and D.~Soudry.
\newblock Accelerated sparse neural training: A provable and efficient method
  to find {N:M} transposable masks.
\newblock 2021.

\bibitem{iofinova2022well}
E.~Iofinova, A.~Peste, M.~Kurtz, and D.~Alistarh.
\newblock How well do sparse imagenet models transfer?
\newblock In {\em Proceedings of the IEEE/CVF Conference on Computer Vision and
  Pattern Recognition}, pages 12266--12276, 2022.

\bibitem{kornblith2019better}
S.~Kornblith, J.~Shlens, and Q.~V. Le.
\newblock Do better imagenet models transfer better?
\newblock In {\em Proceedings of the IEEE/CVF conference on computer vision and
  pattern recognition}, pages 2661--2671, 2019.

\bibitem{kurtic2023ziplm}
E.~Kurtic, E.~Frantar, and D.~Alistarh.
\newblock Ziplm: Hardware-aware structured pruning of language models.
\newblock {\em arXiv preprint arXiv:2302.04089}, 2023.

\bibitem{deepsparse}
M.~Kurtz, J.~Kopinsky, R.~Gelashvili, A.~Matveev, J.~Carr, M.~Goin,
  W.~Leiserson, S.~Moore, B.~Nell, N.~Shavit, and D.~Alistarh.
\newblock Inducing and exploiting activation sparsity for fast inference on
  deep neural networks.
\newblock In H.~D. III and A.~Singh, editors, {\em Proceedings of the 37th
  International Conference on Machine Learning}, volume 119 of {\em Proceedings
  of Machine Learning Research}, pages 5533--5543, Virtual, 13--18 Jul 2020.
  PMLR.

\bibitem{li2021brecq}
Y.~Li, R.~Gong, X.~Tan, Y.~Yang, P.~Hu, Q.~Zhang, F.~Yu, W.~Wang, and S.~Gu.
\newblock {BRECQ}: Pushing the limit of post-training quantization by block
  reconstruction.
\newblock 2021.

\bibitem{liu2022convnet}
Z.~Liu, H.~Mao, C.-Y. Wu, C.~Feichtenhofer, T.~Darrell, and S.~Xie.
\newblock A convnet for the 2020s.
\newblock In {\em Proceedings of the IEEE/CVF Conference on Computer Vision and
  Pattern Recognition}, pages 11976--11986, 2022.

\bibitem{nagel2020up}
M.~Nagel, R.~A. Amjad, M.~Van~Baalen, C.~Louizos, and T.~Blankevoort.
\newblock Up or down? adaptive rounding for post-training quantization.
\newblock In {\em International Conference on Machine Learning}, pages
  7197--7206. PMLR, 2020.

\bibitem{ozkurt2009automatic}
C.~Ozkurt and F.~Camci.
\newblock Automatic traffic density estimation and vehicle classification for
  traffic surveillance systems using neural networks.
\newblock {\em Mathematical and Computational Applications}, 14(3):187--196,
  2009.

\bibitem{beit}
Z.~Peng, L.~Dong, H.~Bao, Q.~Ye, and F.~Wei.
\newblock Beit v2: Masked image modeling with vector-quantized visual
  tokenizers, 2022.

\bibitem{radford2021learning}
A.~Radford, J.~W. Kim, C.~Hallacy, A.~Ramesh, G.~Goh, S.~Agarwal, G.~Sastry,
  A.~Askell, P.~Mishkin, J.~Clark, et~al.
\newblock Learning transferable visual models from natural language
  supervision.
\newblock In {\em International conference on machine learning}, pages
  8748--8763. PMLR, 2021.

\bibitem{ILSVRC15}
O.~Russakovsky, J.~Deng, H.~Su, J.~Krause, S.~Satheesh, S.~Ma, Z.~Huang,
  A.~Karpathy, A.~Khosla, M.~Bernstein, A.~C. Berg, and L.~Fei-Fei.
\newblock {ImageNet Large Scale Visual Recognition Challenge}.
\newblock {\em International Journal of Computer Vision (IJCV)},
  115(3):211--252, 2015.

\bibitem{schuhmann2021laion}
C.~Schuhmann, R.~Vencu, R.~Beaumont, R.~Kaczmarczyk, C.~Mullis, A.~Katta,
  T.~Coombes, J.~Jitsev, and A.~Komatsuzaki.
\newblock Laion-400m: Open dataset of clip-filtered 400 million image-text
  pairs.
\newblock {\em arXiv preprint arXiv:2111.02114}, 2021.

\bibitem{how_to_train_your_vit}
A.~Steiner, A.~Kolesnikov, X.~Zhai, R.~Wightman, J.~Uszkoreit, and L.~Beyer.
\newblock How to train your vit? data, augmentation, and regularization in
  vision transformers, 2021.

\bibitem{DeiT}
H.~Touvron, M.~Cord, M.~Douze, F.~Massa, A.~Sablayrolles, and H.~J{\'e}gou.
\newblock Training data-efficient image transformers \& distillation through
  attention.
\newblock In {\em International Conference on Machine Learning}, pages
  10347--10357. PMLR, 2021.

\bibitem{DeiT3}
H.~Touvron, M.~Cord, and H.~Jégou.
\newblock Deit iii: Revenge of the vit, 2022.

\bibitem{van2018inaturalist}
G.~Van~Horn, O.~Mac~Aodha, Y.~Song, Y.~Cui, C.~Sun, A.~Shepard, H.~Adam,
  P.~Perona, and S.~Belongie.
\newblock The inaturalist species classification and detection dataset.
\newblock In {\em Proceedings of the IEEE conference on computer vision and
  pattern recognition}, pages 8769--8778, 2018.

\bibitem{zafrir2021prune}
O.~Zafrir, A.~Larey, G.~Boudoukh, H.~Shen, and M.~Wasserblat.
\newblock Prune once for all: Sparse pre-trained language models.
\newblock {\em arXiv preprint arXiv:2111.05754}, 2021.

\bibitem{zhang2021self}
L.~Zhang, C.~Bao, and K.~Ma.
\newblock Self-distillation: Towards efficient and compact neural networks.
\newblock {\em IEEE Transactions on Pattern Analysis and Machine Intelligence},
  44(8):4388--4403, 2021.

\bibitem{zhang2019your}
L.~Zhang, J.~Song, A.~Gao, J.~Chen, C.~Bao, and K.~Ma.
\newblock Be your own teacher: Improve the performance of convolutional neural
  networks via self distillation.
\newblock In {\em Proceedings of the IEEE/CVF International Conference on
  Computer Vision}, pages 3713--3722, 2019.

\end{thebibliography}
\bibliographystyle{abbrv}

\newpage

\appendix
\onecolumn

\section{Additional plots}

\subsection{Task-specific vs task-agnostic pruning}\label{app:task_specific_vs_task_agnostic}

Below we present additional plots with a comparison between post-training compression (PTC) and task-aware compression (TACO). 
One can easily observe that task-specific compression works better on subtasks, as stated earlier in our main paper. 
At the same time, we also notice the ``accuracy inversion'' on the generalist ImageNet-1K task due to specialization of our compressed model. 

\begin{figure}[!h]
\centering
    \begin{subfigure}{0.8\linewidth}
        \centering
        \includegraphics[width=\linewidth]{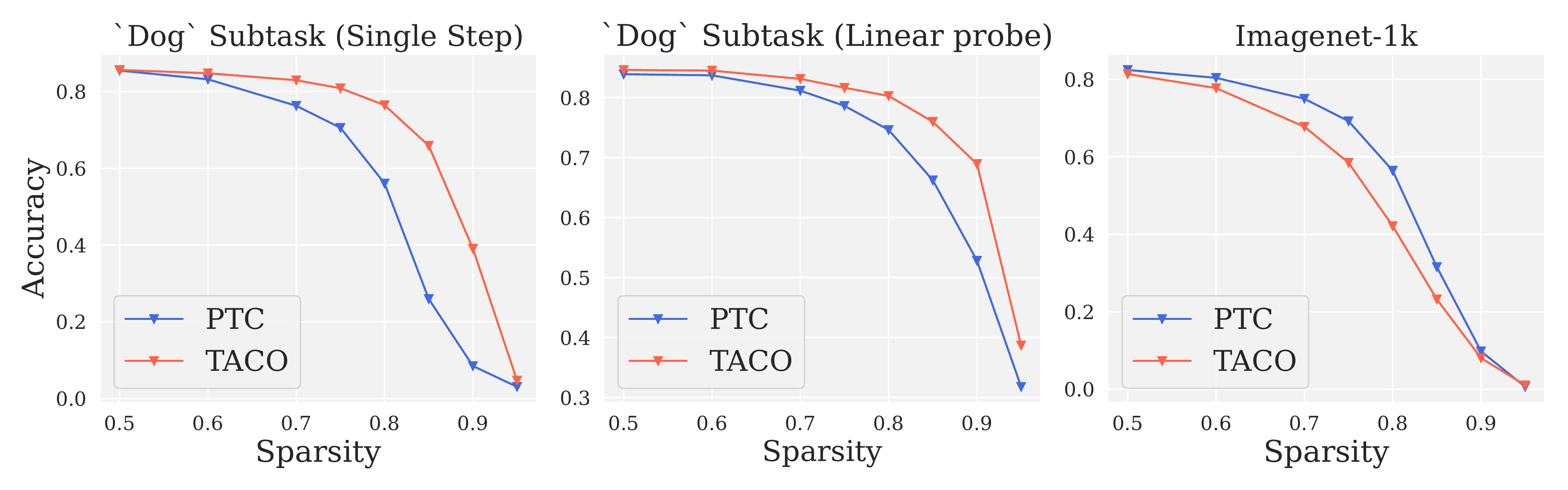}
    \end{subfigure} \\
    \begin{subfigure}{0.8\linewidth}
        \centering
        \includegraphics[width=\linewidth]{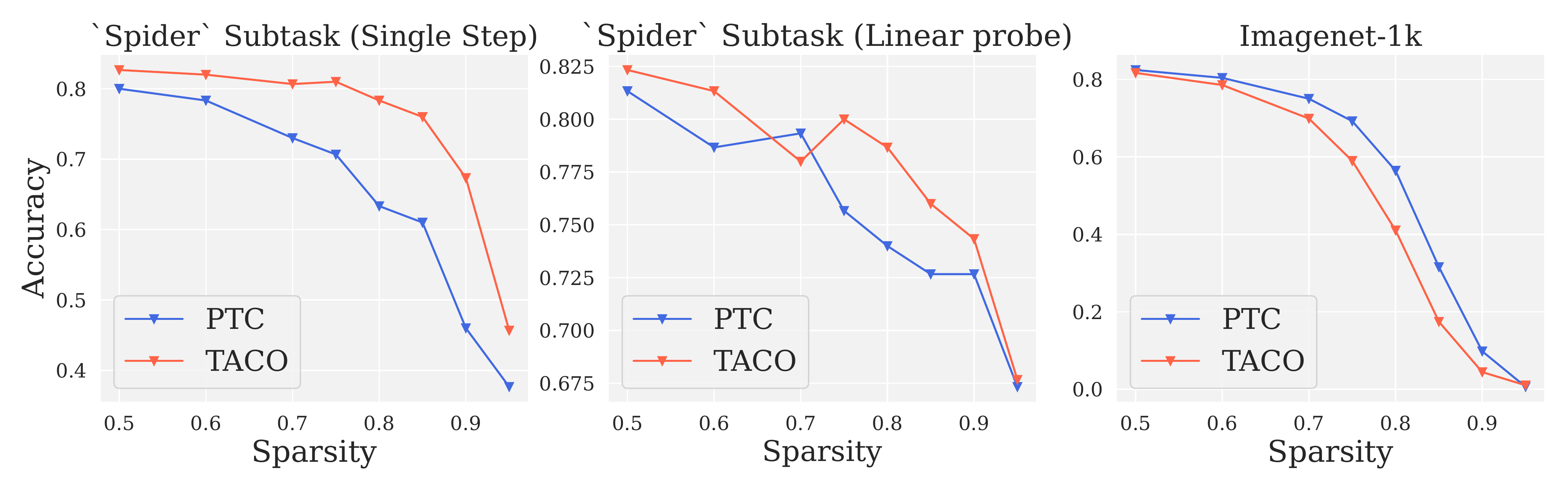}
    \end{subfigure}
    \caption{
        Unstructured pruning of DeiT-III (B/16) model on Dog and Spider subtasks.
    }
    \label{fig:unstructured_pruning_ft_app}
    \vspace{-1em}
\end{figure}

\begin{figure}[!h]
    \centering
    \begin{subfigure}{0.7\linewidth}
        \includegraphics[width=\linewidth]{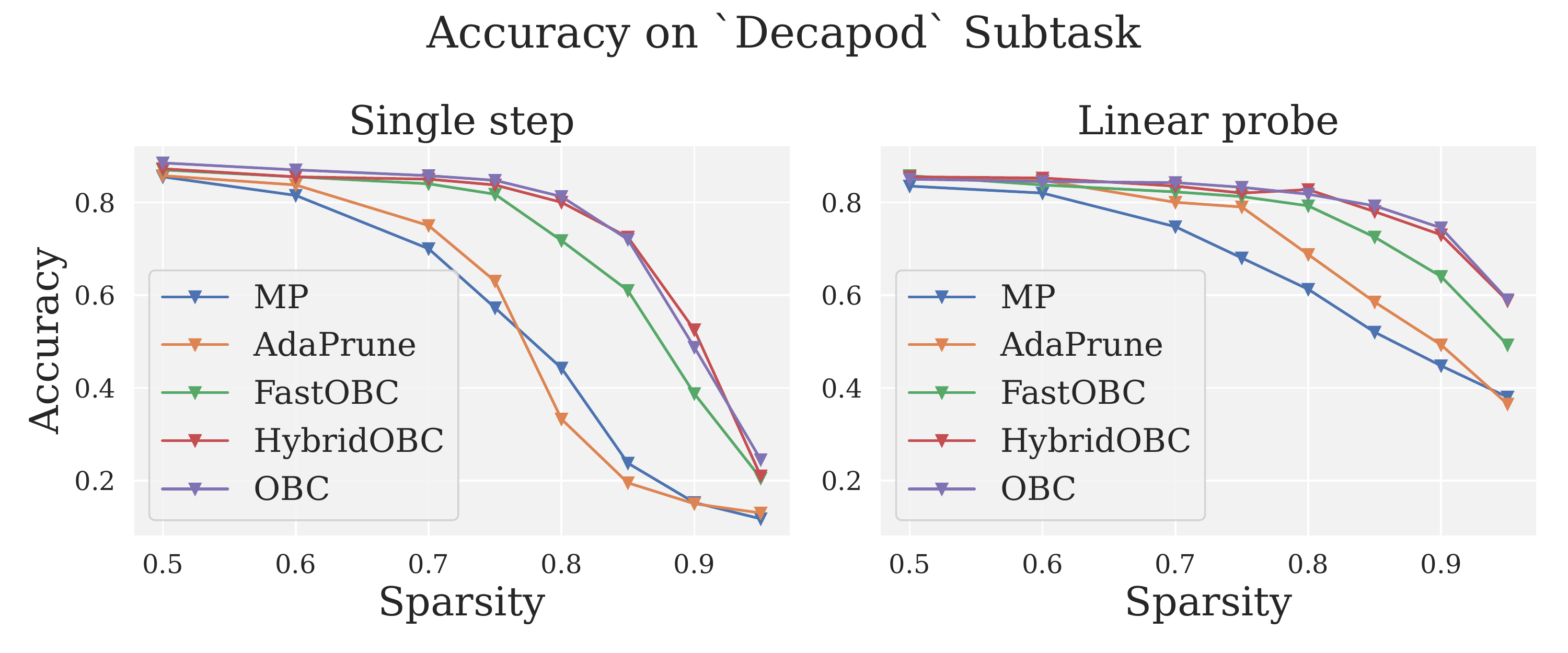}
    \end{subfigure}
    \caption{
         Comparison between sparsity solvers on the DeiT-III (B/16) model.
        (\textbf{Left}) Single step TACO accuracy. 
        (\textbf{Right}) Accuracy after linear probing.
    }
    \label{fig:base_pruner_comparsion_app}
    \vspace{-1em}
\end{figure}

\subsection{Pruner comparison}

Next, we present on Figure \ref{fig:base_pruner_comparsion_app} an additional comparison between sparsity solvers for different sparsity levels. 
One can observe that more advanced and computationally expensive pruners, such as OBC, 
consistently outperform cheaper and faster ones at various sparsities.

\subsection{TACO vs PTC on the ImageNet hierarchy}\label{app:taco_vs_ptc_hierarchy}

\begin{figure}[htb]
    \centering
    \begin{subfigure}{0.25\linewidth}
        \centering
        \includegraphics[width=\linewidth]{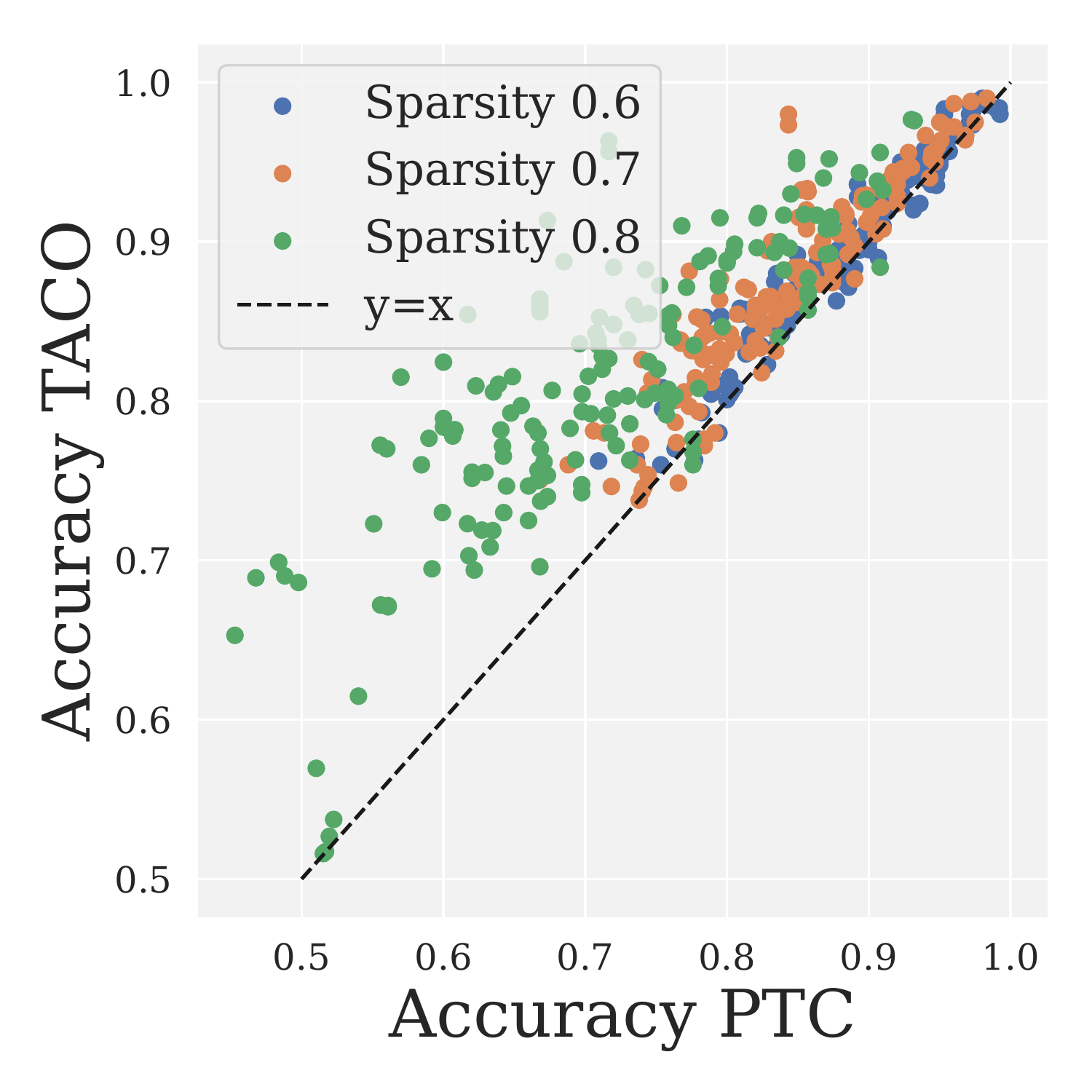}
    \end{subfigure}
    \begin{subfigure}{0.25\linewidth}
        \centering
        \includegraphics[width=\linewidth]{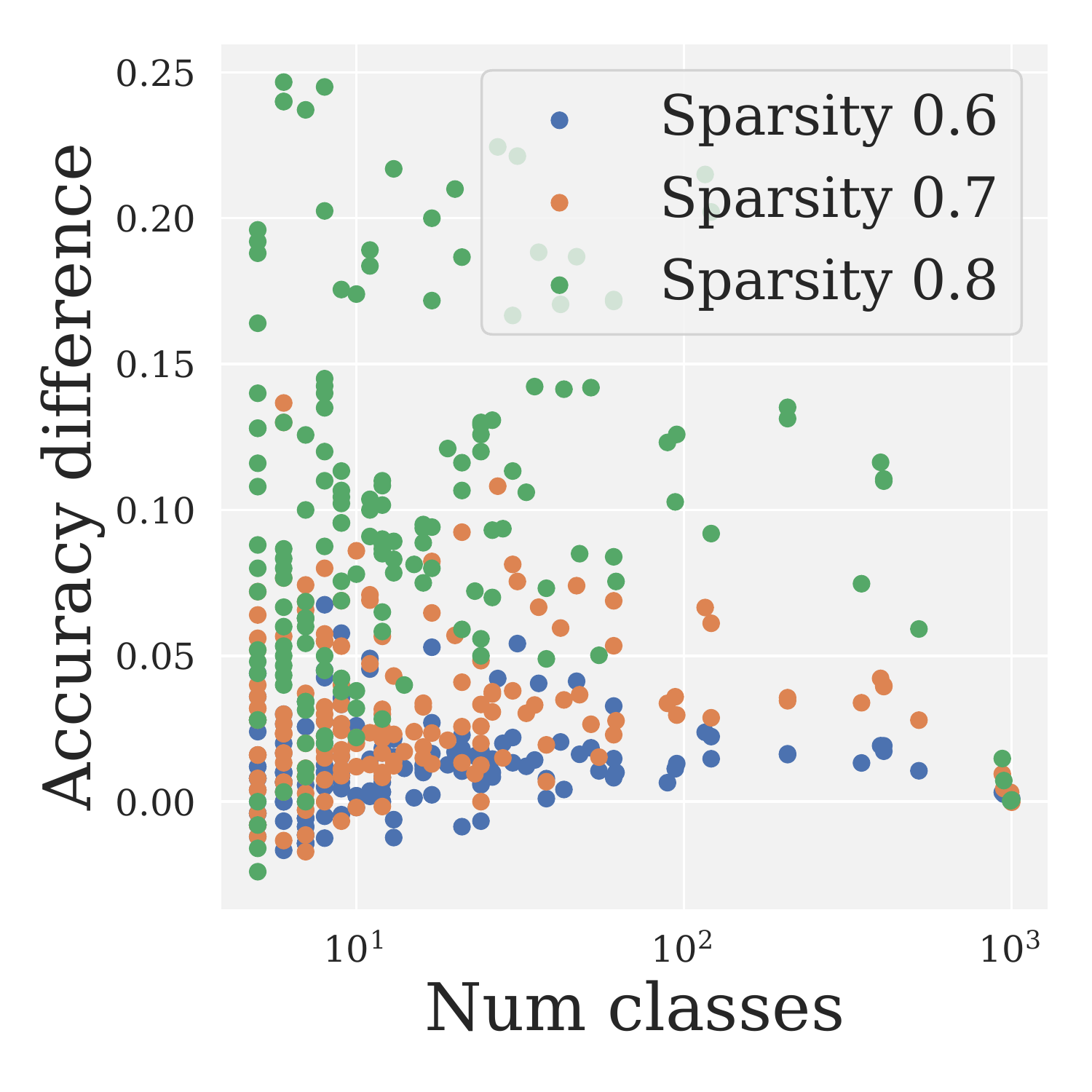}
    \end{subfigure}
    \caption{
       (\textbf{Left}) Accuracy of PTC compressed model vs TACO on the specialized task (ImageNet subset).
       (\textbf{Right}) Accuracy difference between models compressed by TACO and PTC on the specialized tasks.
    }
    \label{fig:compression_hierarchy_taco_vs_ptc}
    \vspace{-1em}
\end{figure}

In order to demonstrate that TACO superiority is not caused by cherrypicking of specific ImageNet subsets 
we evaluated TACO and PTC compression on whole set of ImageNet subtasks with more than 5 classes (i.e 166 subtasks chosen). 
Comparison plots are presented on Figure \ref{fig:compression_hierarchy_taco_vs_ptc}. One can observe
that TACO outperforms PTC on most of the tasks. 

\subsection{Additional plots for iNaturalist}

Below presented is comparison between PTC and TACO on a few more iNaturalist subsets. 

\begin{figure}[!h]
    \begin{subfigure}{0.5\linewidth}
        \centering
        \includegraphics[width=\linewidth]{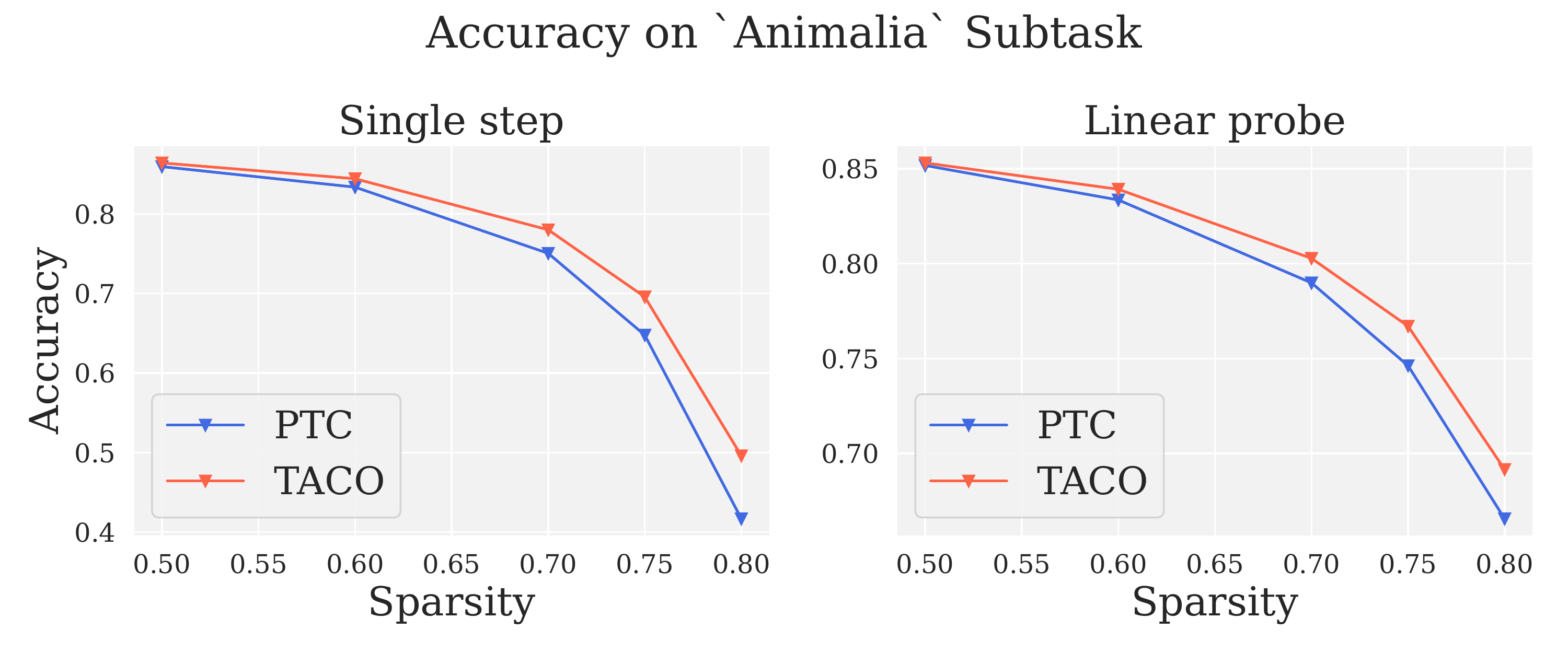}
    \end{subfigure} 
        \begin{subfigure}{0.5\linewidth}
        \centering
        \includegraphics[width=\linewidth]{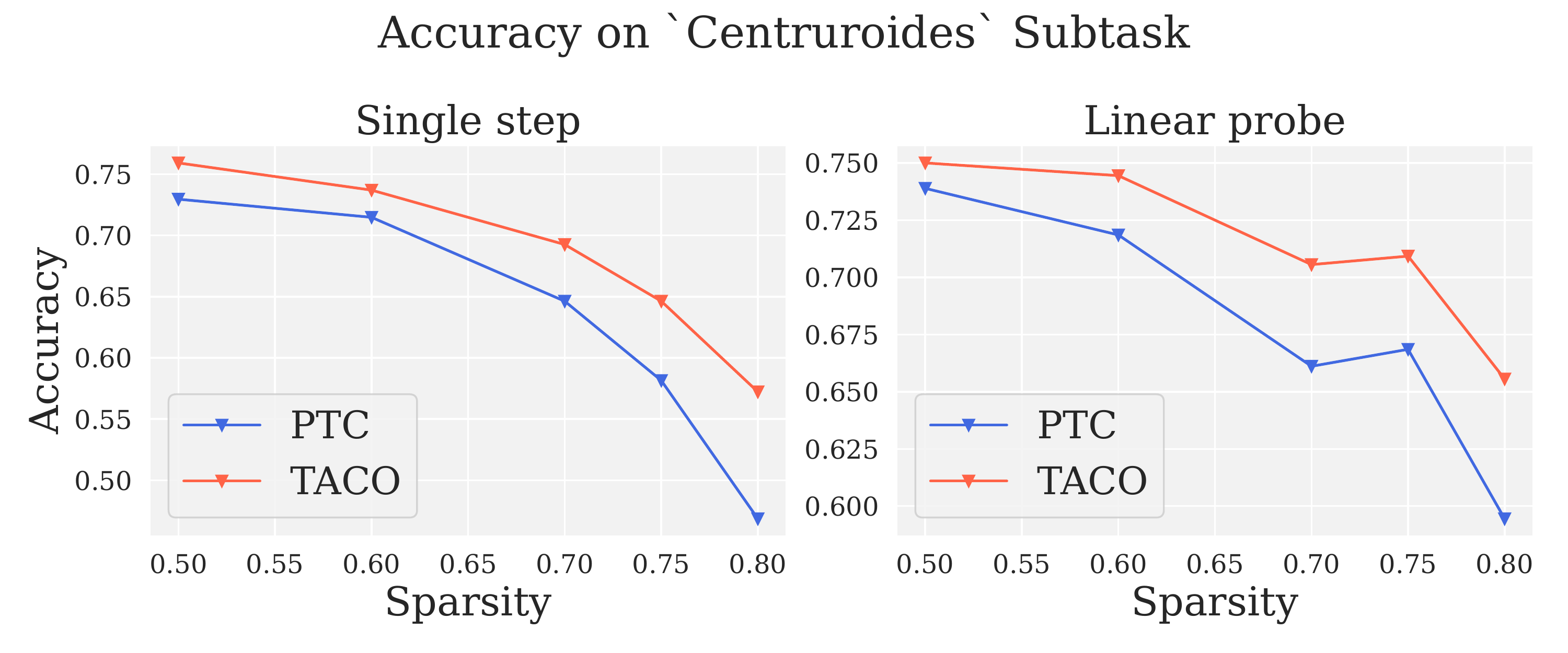}
    \end{subfigure} 
    \centering 
    \begin{subfigure}{0.5\linewidth}
        \centering
        \includegraphics[width=\linewidth]{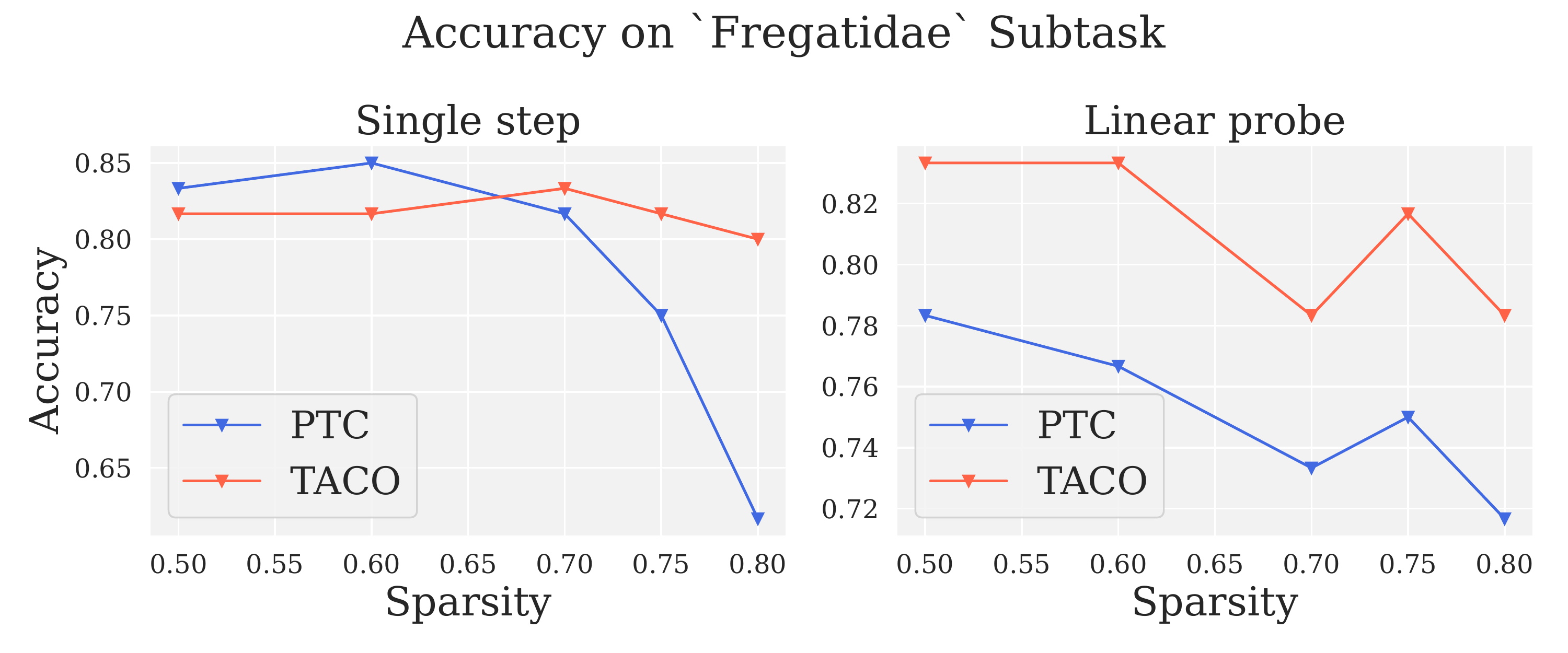}
    \end{subfigure} 
    \caption{
         Unstructured pruning of the DeiT-III (B/16) model on an iNaturalist subtask.
        (\textbf{Left}) Accuracy before finetuning. 
        (\textbf{Right}) Accuracy after linear probing.
    }
    \label{fig:inaturalist_pruning_app}
    \vspace{-1em}
\end{figure}

\section{Finetuning details}\label{app:finetuning_details}

\paragraph{Linear probing} In sparse model finetuning experiments, the linear classifier on top of the frozen backbone is trained for 10 epochs 
using SGD optimizer, batch size 128, learning rate $\eta=0.1$ and momentum $\beta=0.9$. We use the whole subset for 
training the linear classifier.

\paragraph{Quantization-aware training} We train the whole model using SGD optimizer, batch size 128, learning rate $\eta=0.1$ and momentum $\beta=0.9$.
The whole ImageNet subset is used for training. 

\section{TACO Tuner details}\label{app:taco_tuner_details}

In the TACO Tuner setup, a sparsity mask is used to filter the gradients of the pruned weights in every batch. The sparse model is then trained using AdamW, batch size 128, learning rate $\eta=10^{-4}$.

\section{Scaling of compression}

\begin{figure}[htb]
    \centering
    \begin{subfigure}{0.6\linewidth}  
        \includegraphics[width=\linewidth]{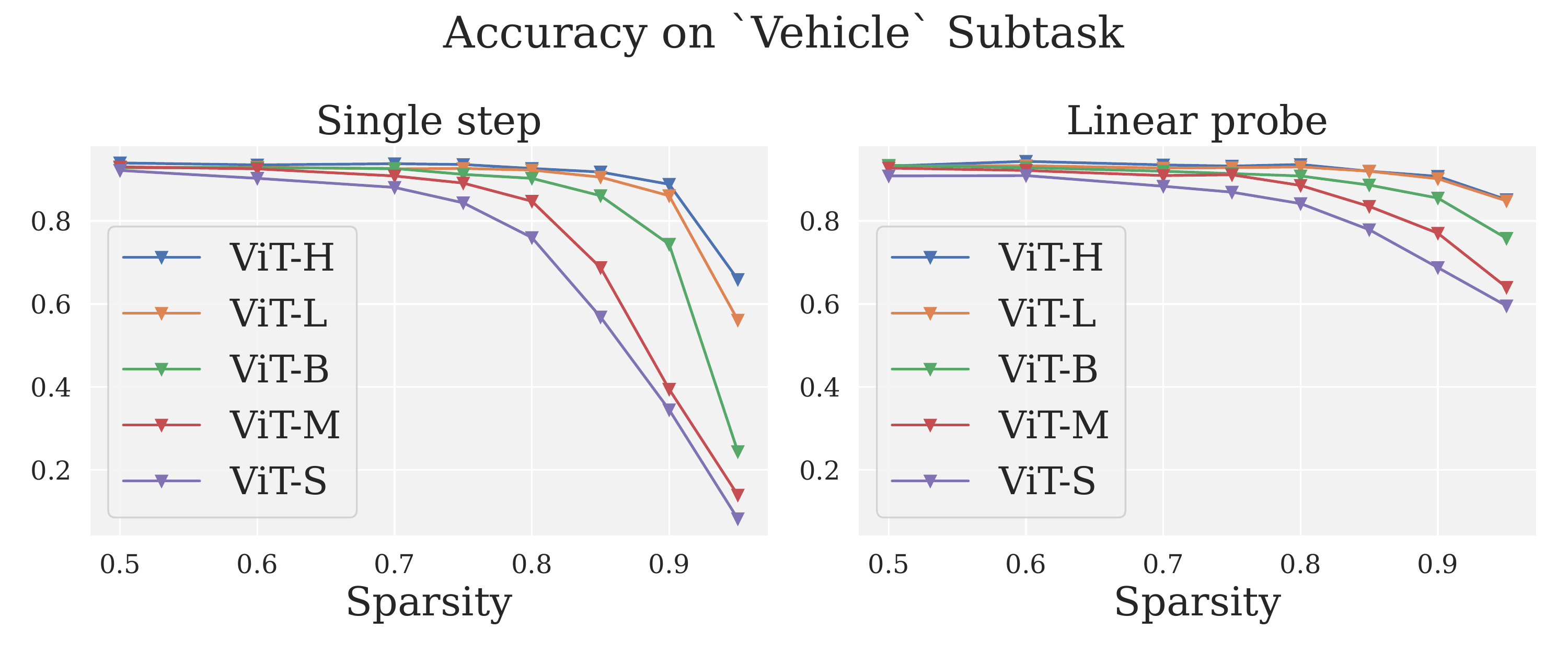}
    \end{subfigure}
    \caption{
        (\textbf{Left}) Accuracy before finetuning. 
        (\textbf{Right}) Accuracy after finetuning.
    }
    \label{fig:scaling_behavior}
\end{figure}

Models from ResNet, ViT, and ConvNeXt families may differ by orders of magnitude in their scale, and one could be 
interested in examining the scaling behavior of the pruning algorithms. Intuitively, larger models being
more over-parametrized should be easier to prune for the same sparsity level. 
We selected 5 models from the DeiT-III family: 
DeiT-III-S/16(224)
, DeiT-III-M/16(224)
, DeiT-III-B/16(224)
, DeiT-III-L/16(224)
and DeiT-III-H/14(224).
And carried out one-shot pruning with HybridOBC (Figure \ref{fig:scaling_behavior}) followed by linear probing.
One can observe that larger ViTs are more compressible compared to smaller models from the same family.

\section{Model comparison}

\begin{figure}[h!]
    \begin{subfigure}{\linewidth}
        \centering
        \includegraphics[width=0.6\linewidth]{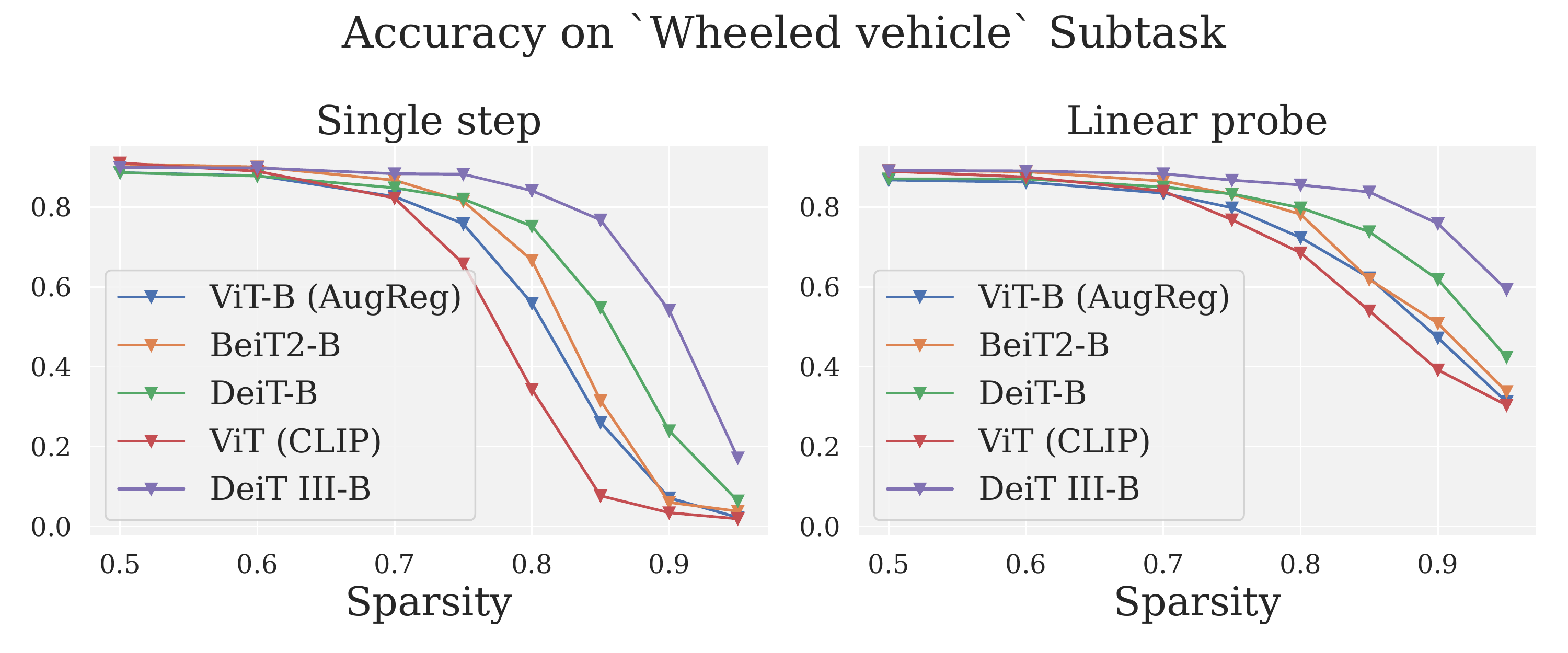}
    \end{subfigure}
    \caption{
        (\textbf{Left}) Accuracy on the whole ImageNet dataset after one-shot pruning. 
        (\textbf{Right}) Accuracy on the subset after one-shot pruning (solid) and 
        linear probing (dashed).
    }
    \label{fig:model_comparison}
\end{figure}

In addition, we have studied the compressibility of different models. 
Interestingly, models of the same architecture with the same number of params can behave very differently 
on an application of pruning methods. One can see from Figure \ref{fig:model_comparison} that 
 DeiT-III-B/16(224) is the most easily compressible, whereas 
AugReg \cite{how_to_train_your_vit}, BEiT v2 \cite{beit} and LAION \cite{schuhmann2021laion}
pretrained ViT CLIP checkpoints drop much faster in accuracy with the increase of sparsity. Our guess is 
that despite the model having same architecture, the resulting weights can be very different, 
and matrices can have a different condition number, eigenvalue distribution, that can 
have a pronounced impact on the performance of the pruning method.

\begin{figure}[h!]
    \begin{subfigure}{\linewidth}
        \centering
        \includegraphics[width=0.4\linewidth]{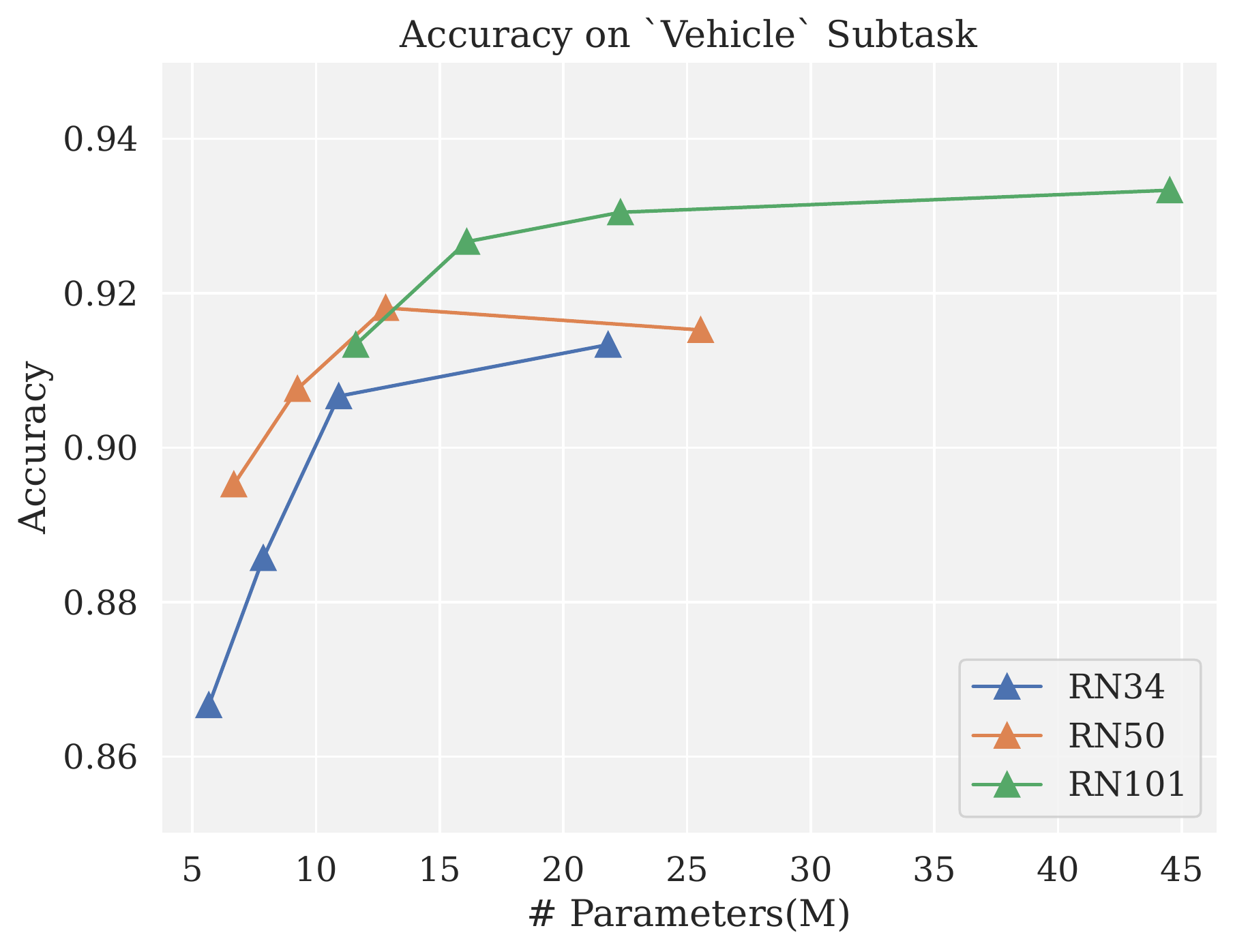}
    \end{subfigure}
    \caption{
        Accuracy on Vehicle task for several TACO-compressed ResNet models. 
        Points on the curves correspond to 50, 65\%, 75\%  sparsity.
    }
    \label{fig:resnet_paramcount}
\end{figure}

A closer look at the parameter efficiency of TACO-produced models is presented in Figure \ref{fig:resnet_paramcount}. 
Variants of ResNet are compressed using TACO + Tuner on the vehicle subtask. 
It can be seen that larger models with higher sparsity attain higher accuracy than smaller models
with lower sparsity.

\end{document}